%% file: main.tex
\documentclass[review,onefignum,onetabnum]{siamonline250211}
\usepackage{booktabs}
\usepackage{tikz-cd}
\usepackage{tikz}

\usepackage{graphicx}
\usepackage{wrapfig}

\input{ex_shared}

\ifpdf
\hypersetup{
  pdftitle={Product Interaction: An Algebraic Formalism for Deep Learning Architectures},
  pdfauthor={
  }
}
\fi

\begin{document}

\maketitle

\begin{abstract}
Modern neural architectures are largely developed as distinct, empirically motivated constructions, despite recurring structural principles across convolutional, attention-based, state-space, and equivariant models. We introduce \emph{Product Interactions}, an algebraic formalism that places these architectures within a common construction principle and turns their design into a systematic algebraic procedure. The central shift is to represent data as elements of an algebra and to regard multiplication, rather than linear transformation alone, as a primitive architectural operation. Architectures are generated by composing multiplication operators of the form
$\mathcal{O}_K(X)=\mathcal{L}_1\!\left(K\mathcal{L}_2(X)\right),$
and their dependence on the input is organized by a notion of \emph{self-interaction order}. Within this formalism, we derive convolutional and equivariant networks as symmetry-constrained first-order product interactions, recover the transition from state-space models to Mamba as an increase in self-interaction order from one to three, and establish attention as a third-order interaction generated by composition of the same underlying operator. We incorporate equivariance through the symmetry principle
$\mathcal{O}\circ\mathcal{T}_g=\mathcal{T}_g\circ\mathcal{O},$
which constrains the algebraic product: translation equivariance recovers convolution, while \(SO(3)\)-compatibility yields the Clebsch--Gordan product and tensor-field interactions.
This perspective recasts architecture design as an algebraic construction problem: the choice of algebra specifies the space of admissible interactions, symmetry constrains their structure, and operator composition systematically generates architectures of different interaction orders. The framework is therefore not only descriptive but generative. Following these principles, we construct a simplified-algebra dynamical layer that improves both accuracy and computational throughput over Mamba on CIFAR-10. Complementary experiments show that interaction order affects accuracy non-monotonically and that symmetry constraints substantially improve the learnability of the underlying algebraic structure. Together, these results establish Product Interactions as a mathematical methodology for analyzing existing neural architectures and systematically constructing new ones from explicit algebraic and symmetry principles.
\end{abstract}



\section{Introduction} 
Early neural network architectures, exemplified by multilayer perceptrons (MLPs)\cite{rosenblatt1958perceptron,rumelhart1986learning}, are structured as compositions of algebraic expressions $\mathcal{W}(X)+b$ and nonlinear activation functions $\sigma$. Building upon this foundation, enhancing the structural expressivity of neural networks fundamentally depends on the design of algebraic expressions that encode inductive biases arising from interaction structures and data symmetries, complemented by appropriate nonlinear activations. A canonical example is the convolutional neural network (CNN)\cite{lecun1998gradient,krizhevsky2012imagenet}, which encodes translation symmetry and thereby fixes the routing of interactions according to this symmetry. Equivariant networks\cite{cohen2016group,cohen2018general} extend this principle to more general group actions, producing symmetry-aware interactions that transform consistently with the data. Attention-based architectures\cite{vaswani2017attention}, by contrast, make the routing of interactions input-dependent. In a related direction, Mamba\cite{gu2023mamba} extends classical state space models (SSMs)\cite{kalman1960new} by making their dynamics input-dependent.

Despite these common structural principles, these architectures are typically formulated and developed as distinct model families, each with its own mathematical construction. Consequently, relationships between their interaction mechanisms are often implicit, and architectural modifications are largely introduced on a model-by-model basis. What is missing is a common mathematical mechanism that can both recover these apparently different constructions and expose explicit design variables along which new architectures can be generated. Such a mechanism would move the study of neural architectures beyond post-hoc comparison toward a systematic procedure in which interaction structure, symmetry, and architectural complexity can be specified and varied explicitly.

In this work, we introduce \textit{Product Interactions}, an algebraic formalism designed to provide such a mechanism. Our starting observation is that several apparently different architectural operations can be organized according to their underlying product structure. Convolutional and equivariant networks implement data--filter interactions constrained by symmetry, whereas Mamba and attention introduce higher-order interactions in which the data itself participates in determining the interaction. This observation motivates a shift from viewing signals only as vectors to viewing them as elements of an algebra, whose product defines the primitive interaction operation.
Within this representation, the algebraic product induces multiplication operators, and architectures are generated through their composition. The construction  separates architecture design into explicit mathematical choices: the algebra determines the admissible product interactions, symmetry constrains their structure, and operator composition controls how these interactions depend on the input. We quantify the latter through the notion of \textit{self-interaction order}. This provides a common construction in which convolutional and equivariant networks arise as symmetry-constrained first-order interactions, the transition from SSMs to Mamba corresponds to increasing the self-interaction order from one to three, and attention arises as a third-order interaction obtained by composition.

Importantly, this perspective is not only descriptive. Once algebra, symmetry, and interaction order become explicit design variables, they can be modified to generate alternative architectures rather than merely reinterpret existing ones. We exploit this generative viewpoint to construct new higher-order product interactions and a simplified-algebra dynamical layer, and use these constructions to investigate how interaction order and algebraic structure affect accuracy and computational efficiency. In this sense, Product Interactions provides a route from structural assumptions to architectural construction: known architectures appear as particular realizations of the formalism, while new architectures arise by changing its algebraic ingredients. To summarize, our main contributions are:

\begin{enumerate}[leftmargin=*, itemsep=1pt]
\item We introduce \textit{Product Interactions}, an algebraic formalism in which neural architectures are generated by embedding data into algebras and composing multiplication operators. This recasts architecture design from model-specific constructions into a common algebraic construction problem, enabling both the analysis and systematic generation of architectures.

\item We introduce \textit{self-interaction order} to characterize how architectures depend on their inputs. Under this notion, convolutional and equivariant networks are first-order interactions, SSMs to Mamba corresponds to the transition from first to third order, and attention arises at third order. Interaction order thereby becomes an explicit mathematical axis for comparing and constructing architectures.

\item We formalize equivariance through $\mathcal{O}\circ\mathcal{T}_g=\mathcal{T}_g\circ\mathcal{O}$ and show that this condition constrains the structure constants of the algebra. Translation symmetry recovers convolution, while $SO(3)$-compatibility yields the Clebsch--Gordan product and tensor-field interactions, turning symmetry into a constructive constraint on architecture rather than a model-specific prescription.

\item We demonstrate that these algebraic degrees of freedom have concrete architectural consequences. Varying interaction order, symmetry constraints, and algebraic structure generates new product interactions, including a simplified-algebra dynamical layer that improves both accuracy and throughput over Mamba on CIFAR-10. \textit{Thus, the formalism provides actionable mathematical principles for neural architecture design.}

\end{enumerate}

\section{Related Work}
\textbf{Existing Algebraic Viewpoints.}
A number of recent works argue that deep learning admits an underlying
algebraic description. 
\cite{gavranovic2024categorical} develop a categorical account of deep
learning architectures via parametric maps and monads, and
\cite{marchetti2024algebra} advocate a neuro\-algebraic–geometric viewpoint
on model families as algebraic varieties.  
\cite{hajij2020aidn} construct \emph{algebraically--informed} networks whose
parameters are constrained to represent given algebraic structures (groups,
Lie algebras, associative algebras), while
\cite{metricconv2024} provide a unifying theory of adaptive convolutions via
metric convolution operators.
These works are primarily concerned with high--level structural or geometric
descriptions of architectures, or with learning algebraic objects
\emph{using} neural networks.
Closer to attention mechanisms, TPAs\cite{zhang2025tensor} improve attention using structured tensor products. Algebraically, attention remains the primitive object in these formulations.
Existing algebraic and geometric view points—whether
categorical,constraint-based,or attention-specific, \textit{ do not provide a mechanism that generates concrete various structurally distinct architectures from a single algebraic operator.}

\textbf{Algebraic and Operator-Theoretic Views of Convolution and Attention.}
Convolutional neural networks were originally introduced to exploit locality and translation symmetry in visual data, beginning with the neocognitron \cite{fukushima1980neocognitron} and later formalised in modern CNN architectures such as LeNet \cite{lecun1998gradient}. Subsequent theoretical work has characterised convolution as a linear equivariant operator under the action of the translation group \cite{mallat2012group}. More recent work studies convolution as an instance of structured linear operators or integral transforms \cite{bronstein2021geometric}.
Attention mechanisms, introduced in neural machine translation~\cite{bahdanau2015neural, luong2015effective} and popularised by the Transformer architecture~\cite{vaswani2017attention}, replace fixed convolutional kernels with data-dependent aggregation operators. Several works aim to unify attention and convolution by expressing both as instances of adaptive or dynamic filtering~\cite{wu2019pay, chen2021xvolution}, or by approximating attention with structured linear operators~\cite{katharopoulos2020transformers, choromanski2021rethinking}. These approaches focus on operator equivalence or approximation within specific architectural families.
By contrast, 
We unify CNNs and attention mechanisms under a common framework of multiplication operators, where CNN filters act as learnable algebraic elements and attention filters manifest as data-driven quadratic products.

\textbf{State-Space Models and Sequence Dynamics.} An alternative line of research revisits classical state-space models (SSMs) as sequence models. The S4 family of models connects continuous-time linear dynamical systems with sequence modeling via structured discretisation schemes~\cite{gu2022hippo, gu2022efficiently}. However, these models are fundamentally limited to linear dynamics in the hidden state.
Mamba~\cite{gu2023mamba} introduces a selective mechanism that allows the input to modulate both state injection and readout, substantially increasing expressivity. 
In our formalism, Mamba naturally arises from SSMs by casting the input as a multiplication filter, and extends to third-order interactions via composition.

\textbf{Group-Equivariant and Representation-Theoretic Networks.} Beyond translation symmetry, equivariant neural networks enforce structured transformations under general group actions \cite{cohen2016group}. 
The most closely related line of work to ours consists of equivariant architectures based on irreducible representations.
Early examples include Harmonic Networks\cite{worrall2017harmonicnetworksdeeptranslation}, which preserve two-dimensional rotational symmetry by leveraging irreducible representations. Tensor Field Networks (TFNs) \cite{thomas2018tensor} and related architectures \cite{kondor2018clebsch, weiler20183d} extend this framework to three-dimensional rotations by modeling features as tensor fields transforming under irreducible representations of $SO(3)$
, with interactions governed by Clebsch–Gordan coefficients. This representation-theoretic approach has been extended to attention-based architectures, including SE(3)-Transformers \cite{fuchs2020se3} and equivariant message-passing networks \cite{batzner20223d}.
Our framework derives these interaction rules by enforcing compatibility between group representations and an algebraic product. Consequently, equivariant convolutions  naturally emerge as product interactions within an algebra.

\section{Product Interaction}
\label{sec:product interaction}
The Product Interaction formalism constructs deep learning architectures through two steps. First, data are embedded into an algebra. Second, architectural operations are constructed by composing multiplication operators on that algebra. We begin by describing the algebraic representation of feature and positional information, before introducing the operators acting on these representations.

\subsection{Embedding data into algebra}
The central idea of our formalism is to treat data as elements of an algebra rather than vectors in a vector space.
We work over a base field $\mathbb{F}\in\{\mathbb{R},\mathbb{C}\}$.  
For completeness:
\begin{definition}[Algebra]
An algebra over $\mathbb{F}$ is a vector space $\mathcal{A}$ equipped with a
bilinear map called the product $\mathcal{A}\times\mathcal{A}\to\mathcal{A}:(a,b) \mapsto ab$, i.e.
$
(a+b)c = ac+bc,\quad a(b+c)=ab+ac,\quad 
(\lambda a)b = \lambda(ab)
$
for all $a,b,c\in\mathcal{A}$ and $\lambda\in\mathbb{F}$.
\end{definition}
We work with algebras constructed as tensor products, equipped with the canonical product induced from each factor. For instance, $(f_1\otimes e_1)(f_2\otimes e_2)=(f_1f_2)\otimes(e_1e_2)$.
A signal $x$ 
is then represented as a tensor 
\begin{equation}
    X =\sum_{\alpha_1,\dots,\alpha_n} X_{\alpha_1,\dots,\alpha_n}\,
e_{\alpha_1}^{(1)}\otimes\cdots\otimes e_{\alpha_n}^{(n)}\in\bigotimes_{j=1}^n\mathcal{A}^{(j)}
\end{equation}
%
This representation allows different aspects of the data to be encoded in different algebraic factors. In particular, we distinguish between feature information, which determines what is being processed, and positional information, which determines how information is organized and exchanged.
\paragraph{Feature Information}
 Feature information is encoded in a feature algebra, which facilitates the construction of interactions through its product structure.
 A typical example is finite dimension algebra $\mathcal{A}$ with 
basis $\{e_i\}$ and product law determined by structure constants
$(\lambda_{ij}^k)$ via
\begin{equation}
    e_i e_j = \sum_k \lambda_{ij}^k e_k,
\end{equation}
\begin{wrapfigure}{r}{0.25\textwidth}
    \centering \vspace{-0.8cm}
    \includegraphics[width=\linewidth]{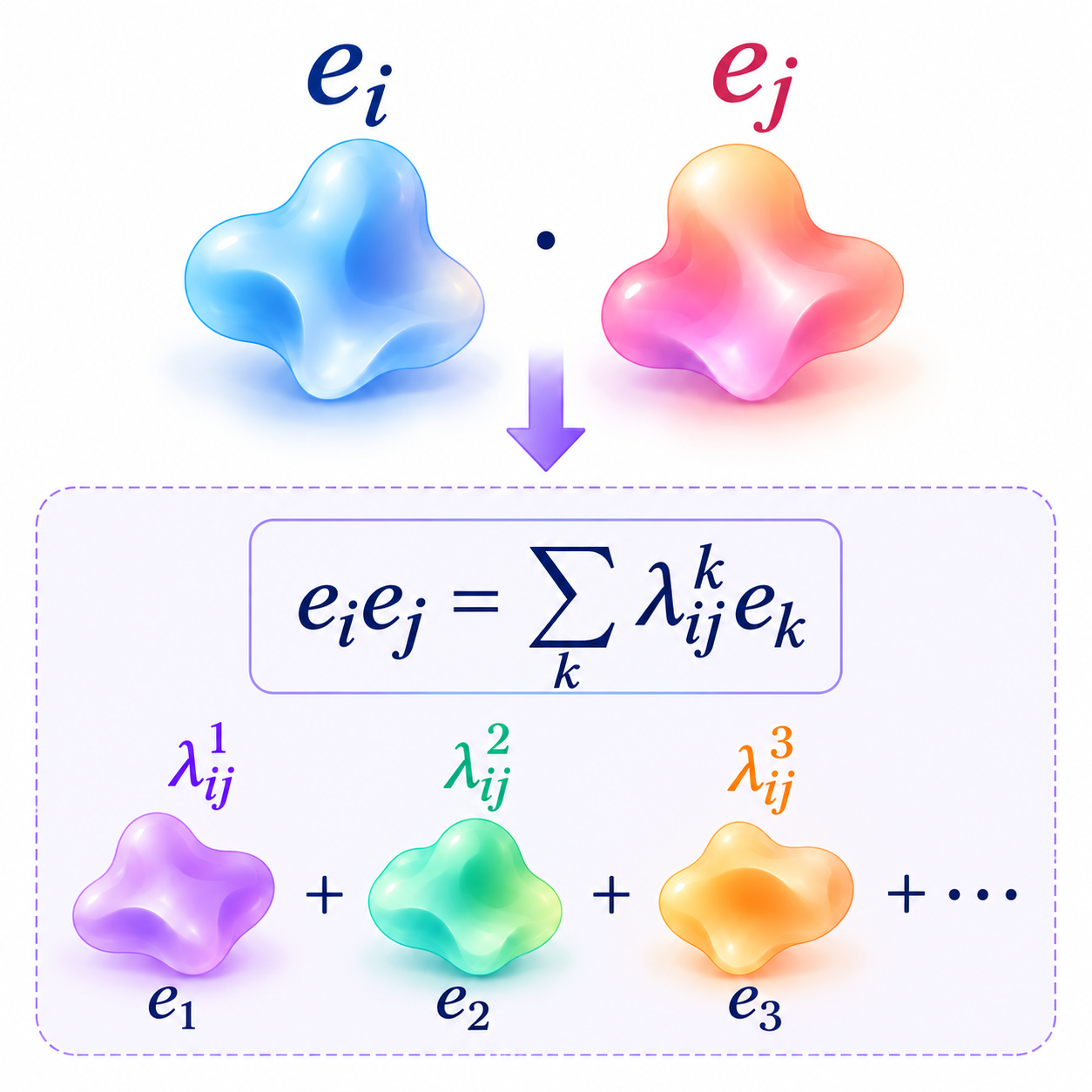}\vspace{-0.7cm}
    \caption{Algebraic multiplication of basis elements, determined by the structure constants $\lambda_{ij}^{k}$.} \vspace{-0.5cm}
    \label{fig:example}
\end{wrapfigure}
which encode all pairwise interactions.  
%
The choice of feature algebra can also incorporate structural properties of the features. In particular, when features transform under a symmetry group, the algebraic product should be compatible with that transformation. For example, for features endowed with rotational symmetry, it is natural to embed the feature information into an algebra compatible with the representation of the rotation group, with its product structure generated by the Clebsch--Gordan coefficients.

\paragraph{Position Information}\label{sec: Position Information}
While the feature algebra determines how feature values interact, positional information determines where and how these interactions take place. Within our formalism, positional information can be incorporated in two ways. The first is through a dynamical formulation, in which positional dependence is represented through temporal evolution. For time-series data whose features are embedded in a feature algebra, we can formulate an algebraic dynamical equation
\begin{equation}
    H(t+1)-H(t)=F(H(t),X(t))
\end{equation}
on the feature algebra to process the data, where \(H(t)\) and \(X(t)\) are elements of the feature algebra, and \(F(H(t),X(t))\) is a product expression defined on the algebra. This is the idea underlying state-space models and Mamba.

The second is to encode position directly through a position algebra. We discretize the space on a lattice, with each lattice point $a$ represented by a basis vector $f_a\in\mathcal{B}$. We then define a product law on this vector space, such that the position-routing mechanism is determined by the algebraic structure of $\mathcal{B}$. For example, convolution can be derived from a product law that respects translation symmetry.

Beyond interactions at individual positions, some architectures require explicit relations between pairs of positions. These can be represented by taking the tensor product $\mathcal{B}\otimes\mathcal{B}$, which allows pairwise positional relations to be treated as algebraic elements. To illustrate this construction, consider $\mathcal{B}=\bigoplus_a F_a,$ where each $F_a\cong\mathbb{F}$ is generated by an idempotent $g_a$ satisfying
\begin{equation}
    g_a g_b=\delta_{ab}g_a,
\qquad
g_0:=\sum_a g_a,
\quad
g_0g_a=g_a.
\end{equation}
Then, \(g_a\otimes g_b\in\mathcal{B}\otimes\mathcal{B}\) can represent an edge between two position nodes. Such an edge can be constructed from two nodes with the help of the flip operator $\mathcal{T}(g_a\otimes g_b)=g_b\otimes g_a,$
as
\begin{equation}
(g_a\otimes g_0)\mathcal{T}(g_b\otimes g_0)
=(g_a\otimes g_0)(g_0\otimes g_b)
=g_a\otimes g_b.
\end{equation}
Once represented algebraically, edges can interact with node information through the same product structure. Specifically, $ (g_a\otimes g_b)\mathcal{T}(g_c\otimes g_0)
=\delta_{bc}g_a\otimes g_b. $
The resulting edge-node interactions can then be aggregated over nodes to update the node information. This aggregation can be implemented by the reduction operator $\mathcal{P}$, defined by $\mathcal{P}(g_a\otimes g_b)=g_a\otimes g_0,$
which gives
\begin{equation}
\mathcal{P}\left[(g_a\otimes g_b)\mathcal{T}(g_c\otimes g_0)\right]
=\delta_{bc}g_a\otimes g_0.
\end{equation}
In practice, we can further introduce a causal projection \(\mathcal{P}^c\) or a neighborhood projection \(\mathcal{P}^N\) to restrict the aggregation to the desired set of nodes. We refer to \(\mathcal{B}\otimes\mathcal{B}\) as the relation algebra in what follows.

\subsection{Composition of multiplication operator}\label{sec: Composition of multiplication operator}
Having represented data as elements of an algebra, we now turn to the second step of the formalism: constructing architectural operations from the algebraic product. Beyond its underlying vector space structure, an algebra provides a multiplication law that induces operators acting on its elements. These operators serve as the basic building blocks of Product Interactions. A single multiplication operator defines an elementary interaction. More complex architectural operations arise by composing these operators, leading to the following definition.

\begin{definition}[Multiplication operator ]
Given structural operators $\mathcal{L}_1$, $\mathcal{L}_2$ and filter element $K\in\bigotimes_{j=1}^n\mathcal{A}^{(j)}$,  the \emph{Multiplication operator}  associated with $(\mathcal{L}_1,\mathcal{L}_2,K)$ is the map
$\mathcal{O}_K :
\bigotimes_{j=1}^n\mathcal{A}^{(j)}\to
\bigotimes_{j=1}^n\mathcal{A}^{(j)}$ defined by
\begin{equation}
  \mathcal{O}_K(X)
    := \mathcal{L}_1(K\mathcal{L}_2(X)),
\end{equation}
\end{definition}
The structural operators in definition of multiplication operator could be flip operator $\mathcal{T}$, reduction operator $\mathcal{P}$ and learnable linear operators. In principle, the multiplication operator depends explicitly on the structural operators \(\mathcal{O}^{\mathcal{L}_1,\mathcal{L}_2}_K\). For simplicity, we denote it by \(\mathcal{O}_K\) throughout, with the structural operators understood from the context.

\begin{definition}[Product interaction ]
    Product interaction operators $\mathcal{O}^{\text{prod}}$ (or \emph{product interactions} for simplicity) are obtained by composing multiplication operators.
\end{definition}
There are two basic ways to compose multiplication operators, depending on whether the output of one interaction is used as the filter or as the input of another. In the first case, the filter is constructed from another interaction:
\begin{equation}\label{eq:first compo}
   \mathcal{O}^{\text{comp}}(X,Y,Z,...)=\mathcal{O}_K(X),\quad K=\mathcal{O}(Y,Z,...).
\end{equation}
In the second case, the output of one interaction becomes the input to a subsequent multiplication operator:
\begin{equation}\label{eq:second compo}
   \mathcal{O}^{\text{comp}}(X,Y,Z,...)=\mathcal{O}_Z(W),\quad W=\mathcal{O}(X,Y,...).
\end{equation}
The distinction is therefore whether composition modifies the filter acting on the input or the input being processed. Both constructions allow interactions to be built recursively from the same underlying multiplication operator.

An important consequence of this construction is that the input may participate in multiple factors of the resulting product expression. In particular, self-interaction can be introduced by identifying previously distinct arguments, as in $\mathcal{O}^{\text{prod}}(X,X,Z,...)$. This motivates a characterization of the resulting dependence on the input.
\begin{definition}[Self-interaction order]
    The Self-interaction order for the input $X$ of the product interaction $\mathcal{O}^{\text{prod}}(X)=\mathcal{O}(X,...,X,Y,Z,...)$ is defined as the polynomial order of  $X$ in expression  $\mathcal{O}(X,...,X,Y,Z,...)$.
\end{definition}
This notion organizes architectural operations according to how many times the input enters their product structure. The following examples illustrate how familiar architectures and their extensions arise through successive compositions.

\textbf{Product interaction with self-interaction order $1$}: Single multiplication operator $\mathcal{O}_K(X)$ provides example of product interaction with self-interaction order $1$. Convolution, gating mechanism, Tensor filed networks and state space models(SSMs) can be regarded as specific realizations of multiplication operator over suitable algebra

\textbf{Product interaction with self-interaction order $2$}:
The easiest way to obtain product interaction with self-interaction order $2$
is taking the filter element $K$ in $\mathcal{O}_K(X)$ to be input $X$
resulting in  quadratic product interaction  $\mathcal{O}^{\text{quad}}(X)= \mathcal{O}_X(X)$. The transition from SSMs to Mamba corresponds precisely to increasing the self-interaction order from 
$1$ to $2$ in this manner.

\textbf{Product interaction with self-interaction order $3$}:
Once we have order $2$ product interaction $\mathcal{O}^{\text{quad}}$, we can composite it with multiplication operator to obtain Product interaction with self-interaction order $3$. In operator form, Transition from CNN to attention corresponds to the composition \eqref{eq:first compo} by taking filter elements $K=\mathcal{O}^{\text{quad}}(X)$ in $\mathcal{O}_K(X)$ resulting in cubic product interaction $\mathcal{O}^{cubic}(X)=\mathcal{O}_{\mathcal{O}_{X}(X)}(X)$.
Gating mechanism of Mamba corresponds to composition \eqref{eq:second compo} by taking filter $Z=X$ serving as gating and input $W=\mathcal{O}^{\text{quad}}(X)$ resulting in 
$\mathcal{O}_{X}(\mathcal{O}_{X}(X))$. 

\textbf{Multi-level product interaction with higher order}:
We can composite product interaction block by block to get more complex product interactions. Construction of SE(3)-attention 
gives an example of such multi-level product interaction which effectively increases self-interaction order of TFN from $1$ to $3$.

Finally we introduce a pointwise nonlinear activation to retain the nonlinear activations used in neural networks
\[
\mathbf{F}(X)=\sum_{J}\mathbf{F}(X_{J})e_J\in \bigotimes_{j=1}^n\mathcal{A}^{(j)},
\]
where $\mathbf{F}$ is nonlinear activation function. 

We emphasize that the interaction order is merely a means of organizing and characterizing architectures. The expressive power does not increase monotonically with the interaction order, but also depends critically on the underlying algebraic structure and the way nonlinear activations are incorporated. This is further discussed in \ref{sec: Increasing self-interaction order}.

\textbf{Notation}: Throughout this paper, symbols in normal font ($W$) denote element in algebra, symbols in calligraphic font ($\mathcal{W}$) denote linear operators, and symbols in bold font ($\mathbf{F}$) denote nonlinear activation functions. So $WX$ should be understand as product of two elements $W$ and $X$ in algebra while $\mathcal{W}(X)$ means apply linear operator $\mathcal{W}$ to element $X$.

\subsection{Symmetry Principle}
The preceding construction generates architectures through algebraic products and compositions of multiplication operators. When the data possess symmetries, these operations must also be compatible with the corresponding transformations. We now show how equivariance can be imposed directly within the Product Interaction formalism.
\begin{wrapfigure}{r}{0.34\textwidth}
    \centering \vspace{-0.5cm}
    \includegraphics[width=\linewidth]{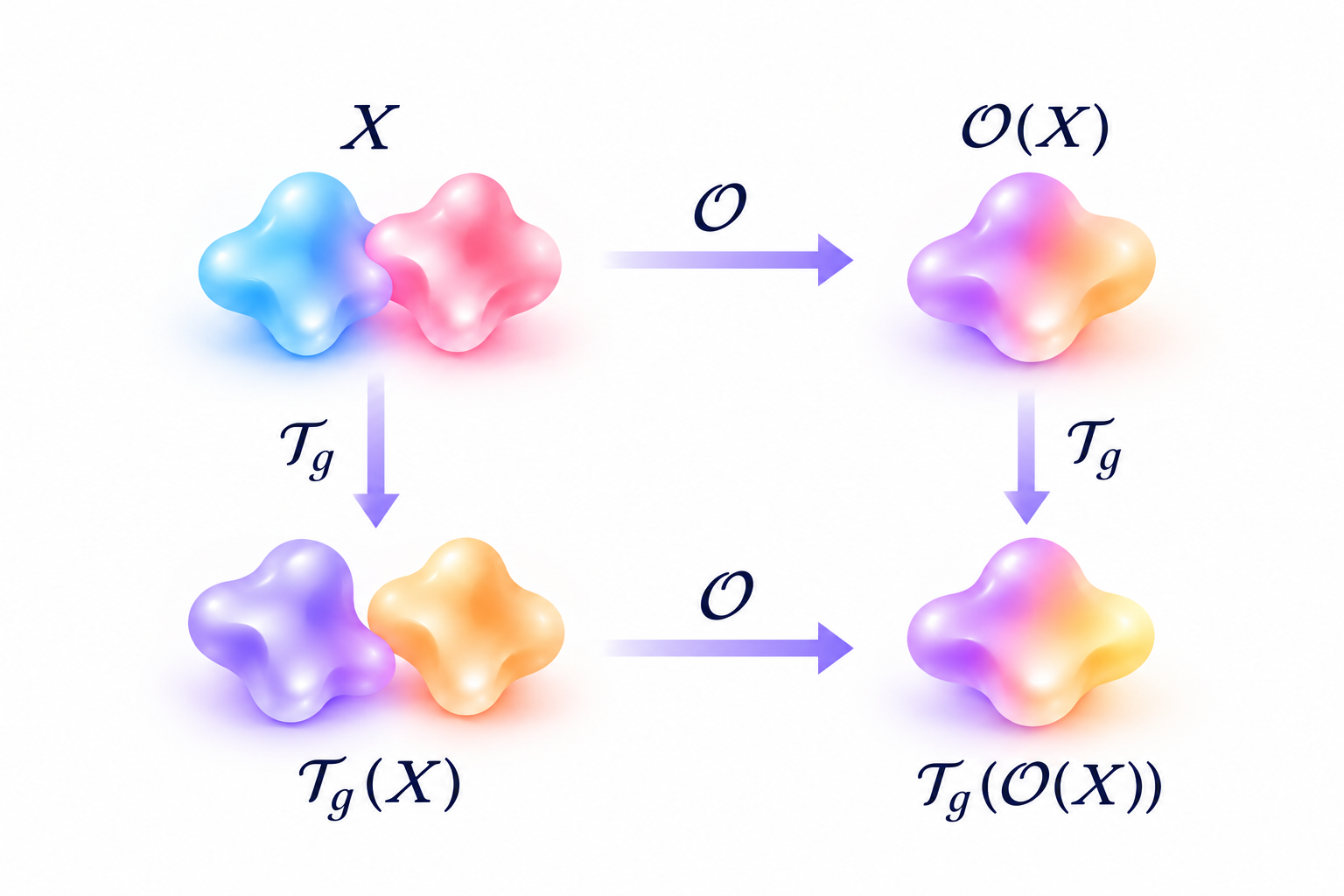}\vspace{-0.7cm}
    \caption{Commutative diagram illustrating the equivariance of a Product Interaction operator under a symmetry transformation.} \vspace{-0.3cm}
    \label{fig:example}
\end{wrapfigure}
After embedding the data into an algebra, a signal is represented by $X\in\bigotimes_{j=1}^n\mathcal{A}^{(j)}$. A symmetry transformation $g$ acting on the original data is represented on the embedded signal by an operator $\mathcal{T}_g$.
\begin{definition}[Symmetry principle]
    For an product interaction operator $\mathcal{O}$, the symmetry equivariant condition is 
    \begin{equation}\label{def:symmetry principle}
\mathcal{O}\circ\mathcal{T}_{g}=\mathcal{T}_{g}\circ\mathcal{O}
    \end{equation}
\end{definition}
The symmetry principle imposes constraints on the structure constants, thereby restricting the form of interactions and facilitating more effective learning.
 We illustrate symmetry principle by deriving CNN as translation equivariant multiplication operator and TFN as SO(3) equivariant multiplication operator.

\section{Product Interaction in Deep Learning}
The product interactions provide a single algebraic
mechanism from which disparate deep–learning architectures arise once one
chooses an appropriate algebra to embed data. In this
section we demonstrate this by explicitly deriving some architectures as realizations of product interaction. 
 We also illustrate how the symmetry principle imposes constraints on the product interaction. 

\subsection{Product interaction with self-interaction order $1$}
We demonstrate that a wide range of network architectures can be viewed as order $1$ multiplication operators defined on appropriate algebras. 

\paragraph{CNNs as symmetry constrained multiplication operator}
\textbf{Step 1: Embed data in Algebra}
We consider a scalar 2D-signal discretized on 2D lattice, and use position algebra $\mathcal{B}$ with basis $\{e_i\}$ to represent the position of the signal. Since we have two axises, we use $\mathcal{B}\otimes\mathcal{B}$ and
embed 2D-signal as
\(X=\sum_{i,j} X_{ij}\, e_i\otimes e_j\).

\textbf{Step 2: Multiplication operator}
We then consider multiplication operator with filter \(K=\sum_{k,l} K_{kl}\, e_k\otimes e_l\),
\begin{equation}\label{eq:CNN-multiplication}
    \mathcal{O}_K(X)=KX
   =\sum_{n,m}
      \Big(\sum_{i,j}\sum_{k,l}K_{kl}X_{ij}\,
 \lambda^{\,n}_{k,i}\lambda^{\,m}_{l,j}\Big)\,
      e_n\otimes e_m,
\end{equation}
where the structure constants $\lambda^{\,n}_{k,i}$ encode horizontal and
vertical interactions.

\textbf{Symmetry principle.} We now impose translation equivariance to determine how the algebraic product is constrained. Under the translation action
$\mathcal{T}_{a,b}(e_i\otimes e_j)=e_{i+a}\otimes e_{j+b}$,
the symmetry principle \eqref{def:symmetry principle} requires the multiplication operator $\mathcal{O}_K$ to commute with $\mathcal{T}_{a,b}$:
\begin{equation}
    \mathcal{O}_K(\mathcal{T}_{a,b}(X))=\mathcal{T}_{a,b}(\mathcal{O}_K(X)),\qquad\forall(a,b)\in\mathbb{Z}^2
\end{equation}
Expanding both sides yields the index–shift constraints
\begin{equation}\label{eq:CNN-shift}
\lambda^{\,n}_{k,i+a}=\lambda^{\,n-a}_{k,i},
\qquad
\lambda^{\,m}_{l,j+b}=\lambda^{\,m-b}_{l,j}.
\end{equation}
The only interesting solution to \eqref{eq:CNN-shift}  is
\(
\lambda^{\,n}_{k,i}=\delta_{k,i-n}
\)
and
\(
\lambda^{\,m}_{l,j}=\delta_{l,j-m}.
\)
Substituting these into $KX$ gives
\begin{equation}\label{eq:exact CNN-multiplication}
  \mathcal{O}_K(X)=  KX
=\sum_{n,m}
   \Big(\sum_{i,j} X_{ij}\, K_{\,i-n,\,j-m}\Big)\,
   e_n\otimes e_m,
\end{equation}
which is precisely the discrete cross–correlation. Consequently, translation symmetry constraint restricts multiplication operator to the form of convolution operator.


\paragraph{Gating mechanism}

\textbf{Step 1: Embed data in Algebra}
    Consider feature algebra $\mathcal{A}$ with product law $e_ie_j=\delta_{ij}e_i$ and two signals $X=\sum_i X_i e_i\in\mathcal{A}$ and $Y=\sum_i Y_i e_i\in\mathcal{A}$.
    
\textbf{Step 2: Multiplication operator}
    The  multiplication operator
\begin{equation}\label{eq:gating}
    \mathcal{O}_{\mathbf{F}(Y)}(X)=\mathbf{F}(Y)X=\sum_i [\mathbf{F}(Y_i)X_i] e_i
\end{equation}
gives gating mechanism for sigmoid nonlinear activation. Composition with gating operator provides an effective way to increase self-interaction order.


\paragraph{Rotation equivariant networks}

\textbf{Step 1: Embed data in Algebra} The essential property of Rotation equivariant networks is symmetry so we consider algebra with representation of rotation group. Let vector space $V$ be direct sum of irreducible $SO(3)$ representations,
$
   V = \bigoplus_{l\in\mathbb{N}} V_l,
   \;
   V_l = \mathrm{span}\{e^{l}_{m}\}_{m=-l,\dots,l},
$
We denote the linear representation  operator as $\rho(R)$ for $R\in SO(3)$.
We now view $V$ as the underlying vector space of the feature algebra
$\mathcal{A}$.  The representation $\rho$ is extended to $\mathcal{A}$, and we
require compatibility of the product with the group action:
\begin{equation}
 (\rho(R)e^{l_1}_{m_1})(\rho(R)e^{l_2}_{m_2})
   = \rho(R)\big(e^{l_1}_{m_1} e^{l_2}_{m_2}\big),
 \quad R\in SO(3).
 \label{eq:so3-compat}
\end{equation}
The product law that satisfy \eqref{eq:so3-compat} is given by standard Clebsch–Gordan
decomposition:
\begin{equation}
   e^{l_1}_{m_1} e^{l_2}_{m_2}
   = \sum_{l}\sum_{m=-l}^{l}
        C^{lm}_{l_1 m_1,\,l_2 m_2}\,e^{l}_{m},
\end{equation}
where $C^{lm}_{l_1 m_1,\,l_2 m_2}$ are Clebsch–Gordan coefficients.

Under irreducible decomposition,  feature field
$s$ can be written as
$
   s(\vec r)
   = \big( s^{l}_{m}(\vec r) \big)_{l,m},
   \;
   s^{l}_{m} : \mathbb{R}^{3}\to\mathbb{C}.
$
The position information is represented  in relation algebra $\mathcal{B}\otimes\mathcal{B}$ introduced in \ref{sec: Position Information}. The final algebra is $\mathcal{B}\otimes\mathcal{B}\otimes\mathcal{A}$.
Let $\{\vec r_a\}_{a=1}^{N}\subset\mathbb{R}^{3}$ be sampling locations,We embed a discretised field
$s=(s^{l}_{m})$ into the algebra $\mathcal{B}\otimes\mathcal{B}\otimes\mathcal{A}$
via
\begin{equation}
   S
   = \sum_{a}\sum_{l,m}
        s^{l}_{m}(\vec r_a)\;
        g_a\otimes g_0\otimes e^{l}_{m},
\end{equation}
Relation algebra allows us to express filter element $K$ as an edge information
\begin{equation}\label{eq:TFN kernel}
   K
   = \sum_{a,b}\sum_{l,m}
        K^{\,l}_{m}(\vec r_a-\vec r_b)\;
        g_a\otimes g_b\otimes e^{l}_{m}.
\end{equation}
 
\textbf{Step 2: Multiplication operator}
In the relation algebra, the interaction between edge and node information is implemented by the flip operator \(\mathcal{T}\otimes\mathrm{Id}_{\mathcal{A}}\), while the aggregation procedure is implemented by the reduction operator \(\mathcal{P}\otimes\mathrm{Id}_{\mathcal{A}}\). For simplicity, we omit \(\mathrm{Id}_{\mathcal{A}}\) and denote these two operators simply by \(\mathcal{T}\) and \(\mathcal{P}\), respectively.
The multiplication operator
$
\mathcal{O}_K(S)=\mathcal{P}\big(K\mathcal{T}(S)\big)
$
then realizes the aggregation of node information through edge relations.
Explicitly,
\begin{equation}\label{eq: multiplication operator for TFN}
    \mathcal{O}_K(S)=\sum_a\sum_{l,m}\{\sum_b\sum_{l_1,m_1}\sum_{l_2,m_2}C_{l_1m_1,l_2m_2}^{lm}K_{m_1}^{l_1}(\vec{r_a}-\vec{r_b})s_{m_2}^{l_2}(\vec{r_b})\}g_a\otimes g_0\otimes e_{m}^{l}
\end{equation}

\textbf{Symmetry principle}
We define transformation of $S$ following geometry definition \eqref{eq:section-action-components} 
    and transfer it to representation of algebra see \ref{app:geometry background for TFN} for details.
\begin{equation*}
     \mathcal{T}_R(S)=\sum_{a}\sum_l\sum_{m'=-l}^ls^l_{m'}(R^{-1}\vec{r_a}) g_a\otimes g_0\otimes (\rho(R^{-1})^*e_{m'}^l) 
\end{equation*}

\emph{Symmetry principle} \eqref{def:symmetry principle}
require that the lifted action $\mathcal{T}_R$ commutes with multiplication operator $\mathcal{O}_K$ 
\begin{equation}
\mathcal{T}_R(\mathcal{O}_K(S))=\mathcal{O}_K(\mathcal{T}_R(S))\qquad R\in SO(3).\label{eq:tfn-equiv-cond}
\end{equation}
This symmetry principle gives same equivariant property of geometry definition \eqref{eq:general symmetry conditon}. 
Kernel satisfy this condition has to take the form(see \ref{app:Symmetry principle for TFN} for details)
\begin{equation}\label{eq: kernal of TFN}
     K^{\,l}_{m}(\vec r)
   = R^{l}(\|\vec r\|)\,Y^{l}_{m}(\hat r),
\end{equation}
with $Y^{l}_{m}$ spherical harmonics and $R^{l}$ an arbitrary radial profile.
Substituting into \eqref{eq: multiplication operator for TFN} recovers the standard Tensor Field
Network \eqref{eq:standard TFN}; the algebraic derivation shows that TFNs are instances of  symmetry constraint multiplication operator.


\paragraph{State space model}
See \ref{app:review mamba} 
for review of SSM.
\textbf{Step 1: Embed data in Algebra}
The principle of SSM is to map inputs to a higher-dimensional hidden state before projecting to the output. In Algebraic formalism this could be realized by considering feature algebra  $\mathcal{A}_0\otimes\mathcal{A}$ where 
$\mathcal{A}_0$ is generated with product law
$f_if_j=\delta_{ij}f_i$ and
$\mathcal{A}$ is generated with product law $e_ie_j=\sum_k\lambda_{ij}^ke_k$. $\mathcal{A}_0$ serves as a hidden algebra and enables the hidden state to be augmented with additional dimensions.

The position information is represented as dynamical equation on this algebra. 
We encode data input $x_\alpha(t)$ and hidden state $h_{\alpha i}(t)$ into
$$X(t)=\sum_{\alpha} x_{\alpha}(t)\, f_{0}\otimes e_{\alpha},
\qquad
H(t)=\sum_{\alpha,i} h_{\alpha i}(t)\, f_{\alpha}\otimes e_{i}.$$

\textbf{Step 2: Multiplication operator}
The flip operator satisfies:
$
\mathcal{T}(f_{0}\otimes e_{\alpha})=f_{\alpha}\otimes e_{0}
$
We take two elements $B=\sum_{\alpha,i}B_{\alpha i}f_\alpha\otimes e_i$, $C=\sum_{\alpha,i}C_{\alpha i}f_\alpha\otimes e_i$ as filters of multiplication operator $\mathcal{O}_B(X)=B\mathcal{T}(X)$ and $\mathcal{O}_C(H)=CH$ respectively. Then consider
the algebraic dynamical system on algebra,
\begin{equation}\label{eq:SSM}
\begin{cases}
\displaystyle 
\frac{dH}{dt}=\mathcal{W}(H) + \mathcal{O}_B(X),\\[0.4em]
\displaystyle 
y_{\alpha}
=\langle \mathcal{O}_C(H),\ f_{\alpha}\otimes e_{0}\rangle.
\end{cases}
\end{equation}
This is exactly dynamical system for SSMs \eqref{eq:classical SSM}, see \ref{app:SSM} 
 for details.

\subsection{Product interaction with self-interaction order $2$}
As mentioned in \ref{sec: Composition of multiplication operator}, order 2 product interaction can be constructed by taking filter element $K$ in $\mathcal{O}_K(X)$ to be $X$ to enhance self-interaction. For example, order 1 gating operator \eqref{eq:gating} is often implemented via $\mathcal{O}_{\mathbf{F}(\mathcal{W}(X))}(X)$. 

Another example is selection mechanism of Mamba. See \ref{app:review mamba} 
for review of Mamba. 
A key design choice in Mamba is to increase the self-interaction order of SSM from 
$1$ to $2$.
We resume the discussion in state space model \eqref{eq:SSM}. SSM use order $1$ multiplication operator $\mathcal{O}_B(X)$ and we can take $B=X$ to get order $2$ product interaction $\mathcal{O}^{\text{quad}}(X)=\mathcal{O}_X(X)$,
\begin{equation}\label{eq:quad of mamba}
    \mathcal{O}^{\text{quad}}(X)=X\mathcal{T}(X)
=\sum_{\alpha,i}
\left(
   \sum_{\beta} \lambda^{i}_{\beta 0} x_{\beta}
\right)
x_{\alpha}\,
f_{\alpha}\otimes e_{i}.
\end{equation}
This shows that input–dependent injection of Mamba is exactly captured by transition from order $1$ multiplication operator to order $2$ product interaction.
Similarly we can take $C=X$ in $\mathcal{O}_C$
\[
\mathcal{O}_X(H)=XH
= \sum_{\alpha,j}
   \Big(
        \sum_{i} h_{\alpha i}
        \sum_{\gamma}\lambda^{j}_{\gamma, i} x_{\gamma}
   \Big)
   f_{\alpha}\otimes e_{j},
\]

Consider the algebraic dynamical system
\begin{equation*}
\begin{cases}
\displaystyle 
\frac{dH}{dt}=\mathcal{W}(H) + \mathcal{O}_X(X),\\[0.4em]
\displaystyle 
y_{\alpha}
=\langle \mathcal{O}_X(H),\ f_{\alpha}\otimes e_{0}\rangle.
\end{cases}
\end{equation*}
This system is exactly Mamba
\eqref{eq:mamba-classical} 
under the identifications
$
W^{B}_{i\beta}=\lambda^{i}_{\beta 0}, 
W^{C}_{i\gamma}=\lambda^{0}_{\gamma, i}.$
Therefore, transition from SSMs to Mamba is exactly increasing self-interaction order by taking filter element of multiplication operator to be input.

\subsection{Product interaction with self-interaction order $3$}

\paragraph{Attention}
    The interaction pattern of attention is determined by input dependent filter.
This principle is naturally captured by increasing self-interaction order of product interaction
$\mathcal{O}_K(X)\xrightarrow{K=\mathcal{O}_X(X)}\mathcal{O}_{\mathcal{O}_X(X)}(X)$.

\textbf{Step 1: Embed data in Algebra}
    To place attention into the algebraic framework, we consider feature algebra
$\mathcal{A}$ defined by structure constant $\lambda_{ij}^k$ and position information is represented by relation algebra $\mathcal{B}\otimes\mathcal{B}$. The final algebra is  $\mathcal{B}\otimes\mathcal{B}\otimes\mathcal{A}$.
Let $x^{(k)}=(x_\alpha^{(k)})\in\mathbb{R}^{d}$ denote the $k$-th token in a sequence.
The input sequence is embedded as
$
   X=\sum_{k,\alpha}
        x^{(k)}_{\alpha}\,
        g_{k}\otimes g_{0}\otimes e_{\alpha}.
$

\textbf{Step 2: Composition of multiplication operator}
Unlike CNNs, where the filter is learned as a set of parameters, attention constructs the filter directly from the data, representing it as an edge relation. Within the relation algebra, this edge relation is naturally obtained by multiplying two node data elements to form an edge element with the help of the flip operator \(\mathcal{T}\otimes\mathrm{Id}_{\mathcal{A}}\) as introduced in \ref{sec: Position Information}. Thus, the data-dependent filter construction in attention is naturally realized as a multiplication operator
\begin{equation*}
    \mathcal{O}_X(X)=X\mathcal{T}(X)=\sum_{k,l}\sum_{\gamma}\big(\sum_{\alpha,\beta}\lambda_{\alpha \beta}^\gamma x^{(k)}_{\alpha}x_\beta^{(l)}\big)g_k\otimes g_l\otimes e_\gamma
\end{equation*}
The information aggregation multiplication operator takes the  form $\mathcal{O}_A(X)=\mathcal{P}(A\mathcal{T}(X))$ where $\mathcal{P}\otimes Id_\mathcal{A}$ is reduction operator.
Then take the quadratic edge element as filter to composite it with aggregation multiplication operator
\begin{equation*}
    \mathcal{O}_A(X)=\mathcal{P}(A\mathcal{T}(X)),\quad A= \mathcal{O}_X(X)=X\mathcal{T}(X) 
\end{equation*}
Expand above equations we get
\begin{equation*}
    \mathcal{O}_{\mathcal{O}_X(X)}(X)=\sum_{k,\theta}\sum_{l}\sum_\gamma
\left(
   \sum_{\alpha,\beta}
   \lambda_{\alpha\beta}^{\gamma}
   x_{\alpha}^{(k)}x_{\beta}^{(l)}
\right)
\left(
   \sum_{\eta}\lambda_{\gamma\eta}^{\theta}
   x_{\eta}^{(l)}
\right)
g_{k}\otimes g_{0}\otimes e_{\theta}.
\end{equation*}

The inner summand
\(
   \sum_{\alpha,\beta}
   \lambda_{\alpha\beta}^{\gamma}
   x_{\alpha}^{(k)}x_{\beta}^{(l)}
\)
acts as an unnormalized interaction score between the query token $k$ and the
key token $l$, while
\(
\sum_{\eta}\lambda_{\gamma\eta}^{\theta}x_{\eta}^{(l)}
\)
plays the role of a value transformation.  


To recover standard attention, we project $A$ to scaler via feature-scalar projection $\mathcal{P}^{0}$ that restricts the output of
the product to the one-dimensional subspace spanned by $e_{0}\in\mathcal{A}$. Then
imposing
$
\lambda_{\alpha\beta}^0=\sum_i\tfrac{1}{\sqrt{d}}W_{i\alpha}^QW_{i\beta}^K
$
ensures that the algebraic score reduces exactly to the dot-product after applying learnable linear maps
$W^{Q},W^{K}$ of attention.

The causality condition can be imposed by causal reduction $\mathcal{P}^{c}(g_{k}\otimes g_{l})=0$ whenever $l>k$ and $\mathcal{P}^{c}(g_{k}\otimes g_{l})=g_k\otimes g_0$ otherwise. In this case, the structural operator $\mathcal{P}$  is replaced by $\mathcal{P}^c$.

Putting these components together, the cubic Product interaction recover attention  is:
\begin{equation}\label{eq:attention cubic}
    \mathcal{O}^{\text{cubic}}(X)=\mathcal{O}_A(X)=\mathcal{P}^c(A\mathcal{T}(X)),\quad  A=\mathbf{F}(\mathcal{O}_X(X))=\mathbf{F}(\mathcal{P}^0(X\mathcal{T}(X)))
\end{equation}
and expanding the above gives
\begin{equation*}
    \sum_{k,\theta}\sum_{l=1}^{k}
\mathbf{F}\!\left\{
\frac{1}{\sqrt{d}}
\sum_{i}
\!\Big(\sum_{\alpha}
    W^{Q}_{i\alpha}x_{\alpha}^{(k)}
\Big)
\!\Big(\sum_{\beta}
    W^{K}_{i\beta}x_{\beta}^{(l)}
\Big)
\right\}
\left(
   \sum_{\eta}
   \lambda^{\theta}_{0\eta}x^{(l)}_{\eta}
\right)
g_{k}\otimes g_{0}\otimes e_{\theta}.
\end{equation*}

When the scalar-projection constants are chosen so that
$\lambda^{\theta}_{0\eta}=W^{V}_{\theta\eta}$, the expression above coincides
exactly with the attention 
including the
causal restriction $l\le k$.  
Thus the cubic product interaction operator $\mathcal{O}^{\text{cubic}}$ reconstructs the full QKV attention
mechanism precisely.  
Attention therefore appears not as an ad hoc construction but as a direct
realization of the fundamental product structure encoded in the algebra
$\mathcal{B}\otimes\mathcal{B}\otimes\mathcal{A}$ and the principle of increasing self-interaction order of product interaction.

Note that attention appears as degenerate case of cubic product interaction as $\mathcal{P}^0$ projects to one-dimensional space. We can maintain more information by applying a rank $R$ projection
$\mathcal{P}^R$.  Empirical results show that this improves performance, confirming our understanding of attention as order $3$ product interaction, see  Table~\ref{tab:copy2}.
We mention that taken feature algebra to be direct sum $\mathcal{A}=\bigoplus_{i=1}^h \mathcal{A}^i$ gives 
multi-head attention.


\paragraph{Discrete rule of Mamba}
 The continuous dynamical system of Mamba uses order $2$ product interaction \eqref{eq:quad of mamba}. The discrete rule of Mamba increases self-interaction order by compositing gating multiplication operator \eqref{eq:gating}  with \eqref{eq:quad of mamba}
resulting in order $3$  product interaction 
$\mathcal{O}_{\mathbf{F}(\mathcal{W}(X))}(\mathcal{O}^{\text{quad}}(X))=\mathcal{O}_{\mathbf{F}(\mathcal{W}(X))}(\mathcal{O}_X(X))$. To be precise, assume there is an unit element in algebra $\mathcal{A}$ as $e_{-1}$  and consider linear operator $\mathcal{W}_g(f_0\otimes e_{\alpha})=\sum_{\beta}\mathcal{W}_{\beta\alpha}f_\beta\otimes e_{-1}$. Apply this linear operator to $X(t)$ gives $$\mathcal{W}_g(X(t))=\sum_{\beta}(\sum_\alpha\mathcal{W}_{\beta\alpha}X_\alpha(t))f_\beta\otimes e_{-1}=\sum_\beta\Delta_\beta(t)f_\beta\otimes e_{-1}$$

Then we can composite multiplication operator for gating \eqref{eq:gating} with the quadratic product interaction operator in \eqref{eq:quad of mamba} to get cubic product interaction operator
\begin{equation*}
    \mathcal{O}_{\mathbf{F}(\mathcal{W}_g(X))}(\mathcal{O}^{\text{quad}}(X))=\mathcal{O}_{\mathbf{F}(\mathcal{W}_g(X))}(X\mathcal{T}(X))=\sum_{\alpha}\Delta_\alpha(t)
\left(
   \sum_{\beta} \lambda^{i}_{\beta 0} x_{\beta}
\right)
x_{\alpha}\,
f_{\alpha}\otimes e_{i}.
\end{equation*}

Similarly $$ \mathcal{O}_{\mathbf{F}(\mathcal{W}_g(X))}(H(t))=\sum_{\alpha,i}\Delta_\alpha(t) h_{\alpha i}(t)\, f_{\alpha}\otimes e_{i}$$

The final algebraic dynamical equation on algebra $\mathcal{A}_0\otimes\mathcal{A}$   for Mamba is
\begin{equation*}
    H(s)-H(s-1)=\mathcal{O}_{\mathbf{F}(\mathcal{W}_g(X(s)))}(\mathcal{W}(H(s-1)))+ \mathcal{O}_{\mathbf{F}(\mathcal{W}_g(X(s)))}(X(s)\mathcal{T}(X(s)))
\end{equation*}


\subsection{Multi-level product interaction with higher order}
\label{subsec:TFN}
We can composite product interactions block by block to get more complex interaction pattern and increase self-interaction order. 
For example, for cubic product interaction $\mathcal{O}_{A}(X ), \, A=\mathcal{O}_X(X)$. If $X$ is further processed by multiplication operator $Y=\mathcal{O}_K(X)$, then the composition could be $\mathcal{O}_{A}(X), \, A= \mathcal{O}_Y(X),\, Y=\mathcal{O}_K(X)$. Product interaction of SE(3)-attention provides example of such multi-level construction which increases order of TFN multiplication operator form $1$ to $3$ 

\textbf{Step 1: Embed data in Algebra} The algebra for SE(3)-attention is the same as TFN which is 
$\mathcal{B}\otimes\mathcal{B}\otimes\mathcal{A}$.
As before, a field is embed as
$ S=\sum_{a}\sum_{l,m}s^{l}_{m}(\vec r_a)\,
      g_a\otimes g_0\otimes e^{l}_{m},$
and filter element as
$ W=\sum_{a,b}\sum_{l,m}W^{\,l}_{m}(\vec r_a-\vec r_b)\,
      g_a\otimes g_b\otimes e^{l}_{m}.$
We need two filter elements $W^K$ and $W^V$ to get $K$ and $V$ respectively and they take the same form as TFN \eqref{eq: kernal of TFN}.

\textbf{Step 2: Composition of multiplication operator}
To construct SE(3) attention,
The fist level is to get $K$ and $V$ via multiplication operator $\mathcal{O}_W(S)=W\mathcal{T}(S)$ and $K=\mathcal{O}_{W^K}(S)$, $V=\mathcal{O}_{W^V}(S)$. Note that this is edge-node interaction which gives an edge information.
In SE(3)-attention, $Q$ is obtained from $S$ by linear operator that only acts on channel dimension.
Since we do not consider channel dimension here, $Q$ is simply taken to be $Q=S$. The second level is constructing data dependent filter through multiplication operator $A=\mathbf{F}(\mathcal{O}_Q(K))=\mathbf{F}(\mathcal{P}^0(QK))$ where $\mathcal{P}^{0}$ is scalar projection acts on $\mathcal{A}$, projecting onto
the one–dimensional subspace spanned by $e^{0}_{0}$ (the scalar irrep). The third level is information aggregation through
$\mathcal{O}_A(V)=\mathcal{P}(AV)$ where $\mathcal{P}$ is reduction operator. In practice, the reduction operator is replaced by 
neighbourhood reduction $\mathcal{P}^{N}$ that acts on
$\mathcal{B}\otimes\mathcal{B}$ by $\mathcal{P}^{N}(g_a\otimes g_b)=g_a\otimes g_0,  \vec r_b\in\mathcal{N}_a,$ and $\mathcal{P}^{N}(g_a\otimes g_b)=0$ otherwise
where $\mathcal{N}_a$ is a prescribed neighbourhood of $\vec r_a$.

The overall Multi-level product construction for SE(3)-attention is the following
\begin{align}\label{eq:multi SE(3)}
    &\mathcal{O}^{\text{multi}}(S)=\mathcal{O}_A(V)=\mathcal{P}(AV), \quad A=\mathbf{F}(\mathcal{O}_Q(K))=\mathbf{F}(\mathcal{P}^0(QK))\\
    & K=\mathcal{O}_{W^K}(S)=W^K
    \mathcal{T}(S), \quad V=\mathcal{O}_{W^V}(S)=W^V
    \mathcal{T}(S), \quad Q=S
\end{align}
This recovers standard SE(3)-attention \eqref{eq:standrad se(3) attention}, see \ref{app:SE(3) attention} for details.

\textbf{Symmetry principle} We verify \emph{symmetry principle} \eqref{def:symmetry principle} of multi-level product interaction operator $\mathcal{O}^{\text{multi}}$ \eqref{eq:multi SE(3)}. 
After transformation $S\to \mathcal{T}_R(S)$ we have $Q,K,V\to \mathcal{T}_R(Q,K,V)$ where we have used definition \eqref{eq: transformation of edge}. Note that $\mathcal{T}_R(Q)\mathcal{T}_R(K)=\mathcal{T}_R(QK)$ since the product is consistent with representation \eqref{eq:so3-compat}. For projection to irreducible representation we have $\mathcal{P}^0\circ\mathcal{T}_R=\mathcal{T}_R\circ\mathcal{P}^0$, so $A \to \mathcal{T}_R(A)$. Then $AV\to\mathcal{T}_R(A)\mathcal{T}_R(V)=\mathcal{T}_R(AV)$ also by consistent condition \eqref{eq:so3-compat}.  Finally since $\mathcal{T}_R\circ\mathcal{P}=\mathcal{P}\circ\mathcal{T}_R$(see \ref{app: se(3) symmetry}), we get $\mathcal{O}^{\text{multi}}(\mathcal{T}_R(S))=\mathcal{T}_R(\mathcal{O}^{\text{multi}}(S))$.

\section{Structure of Product Interactions}
The Product Interaction formalism provides a common algebraic representation of several neural architectures. Beyond this unifying description, it also identifies structural choices that govern the construction of new architectures. We examine three such choices: self-interaction order, the underlying algebra, and the structural operators used to compose multiplication operators.

\subsection{Increasing self-interaction order}
\label{sec: Increasing self-interaction order}
By constructing network layers as product interactions, we observe a clear evolution of network architectures toward higher-order interactions that enhance self-interaction. For example, in operator form, the transition from CNN to attention can be expressed as
$\mathcal{O}_K(X)
\xrightarrow{K=\mathcal{O}_X(X)}
\mathcal{O}_{\mathcal{O}_X(X)}(X).$
Similarly, the transition from TFN to SE(3) attention can be written as
\begin{equation}
\mathcal{O}^{\mathrm{TFN}}_K(S)
=\mathcal{P}\big(K\mathcal{T}(S)\big)
\longrightarrow
\mathcal{O}^{\mathrm{Attn}}_A(V)
=\mathcal{P}\big(AW^V\mathcal{T}(S)\big)
=\mathcal{O}^{\mathrm{TFN}}_{A W^V}(S).
\end{equation}
In deriving the above equation, we use the associativity of the algebra \(\mathcal{B}\otimes\mathcal{B}\otimes\mathcal{A}\). Thus, the transition from TFN to attention can be viewed as replacing the learned kernel \(K\) with a data-dependent kernel \(A W^V\), where \(A\) is constructed directly from the input data.
And transition from SSMs to Mamba follows a process of increasing self-interaction order in the following way
\begin{align}
&\mathcal{O}_B(X)\xrightarrow{B=X}\mathcal{O}_{X}(X)\xrightarrow{\mathcal{O}_{\mathbf{F}(\mathcal{W}(X))}(\cdot)}\mathcal{O}_{\mathbf{F}(\mathcal{W}(X))}(\mathcal{O}_X(X))
\end{align}

The above processes have been studied extensively from diverse perspectives. In contrast, our formalism provides a unified perspective on the progression toward higher-order interactions through algebraic products. Therefore, product interactions provide a principled framework for constructing interactions with arbitrary self-interaction orders.


We next demonstrate how to follow the principle of increasing interaction order in the product interaction framework to construct a model that achieves both efficiency and strong performance.
Our starting point is the dynamical formulation of Mamba:
\begin{equation}\label{eq: dyanicaml equation}
    \begin{cases}
            H(s)-H(s-1)=\mathcal{O}_{\mathbf{F}(\mathcal{W}_g(X(s)))}(\mathcal{W}(H(s-1)))+ I(s)\\
            Y(s)=\mathcal{O}_{X(s)}(H(s))
    \end{cases}
\end{equation}
In Mamba, the dynamical equation \eqref{eq: dyanicaml equation} is formulated on the algebra \(\mathcal{A}_0\otimes\mathcal{A}\), and the final expression for \(I\) takes the form of a third-order interaction  $I(s)=\mathcal{O}_{\mathbf{F}(\mathcal{W}_g(X(s)))}\big(\mathcal{O}_{X(s)}(X(s))\big)$.

Note that the dynamical equation is an abstract formulation that can be defined on any algebra. Therefore, we need not restrict ourselves to the original formulation of Mamba and are free to choose other algebras. We then consider a simpler algebra \(\mathcal{A}\) and construct \(I\) according to the principle of increasing interaction order through multiplication operators.

Explicitly, algebra $\mathcal{A}$ is generated with basis $\{e_i,f_{i}\,; i=1,...,d\}$ and product law
\begin{equation}
    e_ie_j=\lambda_{ij}^ke_k, \qquad f_je_i=\delta_{ij}e_i
\end{equation}
We do not need other product relations so we leave them undefined. Data and hidden state are represented as  $X(t)=\sum_i x_i(t)e_i,\quad H(t)=\sum_ih_i(t)e_i$.
Linear operator $\mathcal{W}_g$ is defined to map $e_i$ to $f_j$, $\mathcal{W}_g(e_i)=\sum_j W^g_{ji}f_j$, then 
\begin{equation}
\mathbf{F}(\mathcal{W}_g(X))=\sum_j\mathbf{F}(\sum_iW^g_{ji}x_i(t))f_j=\sum_j\Delta_j(t)f_j,\quad  \mathcal{O}_{\mathbf{F}(\mathcal{W}_g(X))}(H)=\sum_i \Delta_i(t)h_i(t)e_i.
\end{equation}
The output $Y$ is $Y=\mathcal{O}_{X}(H)=\sum_k\big(\sum_{ij}\lambda_{ij}^kx_ih_j\big)e_k$.
We than construct $I$ by iteratively increasing interaction order form 1 to 3.
\begin{equation}
I=X\xrightarrow{\mathcal{O}_X(\cdot)}I=\mathcal{O}_X(X)\xrightarrow{ \mathcal{O}_{\mathbf{F}(\mathcal{W}_g(X))}(\cdot)}I=\mathcal{O}_{\mathbf{F}(\mathcal{W}_g(X))}(\mathcal{O}_X(X))
\end{equation}
where
\begin{equation}
\mathcal{O}_X(X)=\sum_k(\sum_{i,j}\lambda_{ij}^kx_ix_j)e_k,\quad   \mathcal{O}_{\mathbf{F}(\mathcal{W}_g(X))}(\mathcal{O}_X(X))=\sum_k\Delta_k(\sum_{i,j}\lambda_{ij}^kx_ix_j)e_k
\end{equation}
We could also use multiplication operator to increase order of $H$ via
\begin{equation}
    H\xrightarrow{\mathcal{O}_{\mathbf{F}(\mathcal{W}_h(H))}(\cdot)}\mathcal{O}_{\mathbf{F}(\mathcal{W}_h(H))}(H)=\sum_j\big(\mathbf{F}(\sum_i W^h_{ji}h_i)h_j\big)e_j
\end{equation}
where $\mathcal{W}_h(e_i)=\sum_j W^h_{ji}f_j$. The final output is then
$Y=\mathcal{O}_{X}(\mathcal{O}_{\mathbf{F}(\mathcal{W}_h(H))}(H)).$ 


\paragraph{Numerical Experiments.} We investigate the effect of self-interaction order and the choice of algebra through a numerical comparison on CIFAR-10. Using Vision Mamba as the baseline architecture, we replace its Mamba dynamical layer with the proposed constructions. The resulting models are distinguished by the self-interaction orders of (I) and (H), allowing us to examine their effects on classification accuracy. We also compare training and inference throughput to assess the computational consequences of the simplified algebra. The results are reported in Table~\ref{tab: lambdalambda}.

\begin{table}[t!]
\centering
\caption{Comparison of classification accuracy, training throughput, and inference throughput for Vision Mamba and the proposed dynamical layers on CIFAR-10. The proposed models are indexed by the self-interaction orders of the input term $I$ and hidden state $H$. Accuracy is reported as mean $\pm$ standard deviation, and throughput is measured in iterations per second. Furthermore, as speed differences across models of different orders are negligible, we report the performance of the highest-order model, with omitted values indicated by (–).}
\label{tab: lambdalambda}
\small
\resizebox{\textwidth}{!}{%
\begin{tabular}{lccccc}
\toprule
\textbf{Model} & Mamba & $(I=3,H=2)$ & $(I=3,H=1)$   & $(I=2,H=1)$   & $(I=1,H=1)$ \\
\midrule
Accuracy  & $86.818 \pm 0.141$  & $87.118 \pm 0.254$ & $86.748 \pm   
0.192$ & $85.760 \pm
0.152$ & $86.112\pm
0.218$  \\
\midrule
Training speed & 4.81 it/s & 12.95 it/s &    --& -- & --\\ 
\midrule
Inference speed &15 it/s & 39 it/s  & -- & -- & --   \\
\bottomrule
\end{tabular}
}
\end{table}
\begin{table}[t!]
\small
\centering
\caption{MNIST classification accuracy for multiplication operators with exact symmetry constraints (With Sym), without symmetry constraints (Without Sym), and with symmetry constraints imposed through regularization (Reg Sym).}
\label{tab:MNIST}
\begin{tabular}{lccc}
\toprule
\textbf{Model} & With Sym & Without Sym  & Reg Sym \\
\midrule
Accuracy  & 0.988 & 0.445 & 0.943 \\
\bottomrule
\end{tabular}
\end{table}

We observe that increasing the interaction order of \(I\) from 1 to 3 generally improves the accuracy, from \(86.112\) to \(86.748\). However, the improvement is not monotonic, indicating that there is no simple monotonic relationship between interaction order and model performance. Nevertheless, the results demonstrate that increasing the interaction order provides an effective principle for systematically constructing more expressive architectures. Furthermore, increasing the interaction order of \(H\) from 1 to 2 improves the accuracy from \(86.748\) to \(87.118\), further supporting the effectiveness of higher-order product interactions.

\subsection{Algebraic structures} The choice of algebraic structure determines which interactions can be represented and how they are constrained. In addition to self-interaction order, three aspects are particularly relevant to architectural design: compatibility with symmetry, the choice of the underlying algebra, and the structural operators used in constructing multiplication operators.
\paragraph{Symmetry principle}
The symmetry principle~\eqref{def:symmetry principle} constrains the algebraic product through its compatibility with a prescribed symmetry transformation. In the constructions of CNNs and TFNs, the symmetry principle determines the form of the corresponding equivariant interactions. It therefore provides a systematic way to restrict the admissible structure constants, rather than treating them as unconstrained learnable parameters.

\paragraph{Numerical evaluation of symmetry constraints} The effect of symmetry constraints on learning is examined through an MNIST classification experiment. We evaluate the multiplication operator $\mathcal{O}_K$ \eqref{eq:CNN-multiplication} under three settings: exact symmetry constraints corresponding to \eqref{eq:exact CNN-multiplication}, no symmetry constraints, and symmetry constraints \eqref{eq:CNN-shift} enforced via a regularization term $L_{\text{reg}}=\sum_{k,i,n,a}(\lambda^{\,n}_{k,i+a}-\lambda^{\,n-a}_{k,i})^2$ in the loss.

Table~\ref{tab:MNIST} shows that the unconstrained model achieves substantially lower accuracy than either symmetry-constrained formulation. Exact enforcement gives the highest accuracy, while regularization toward the symmetry constraints also yields a considerable improvement. These results suggest that symmetry constraints facilitate learning by restricting the admissible structure constants $\lambda_{ij}^k$.

\paragraph{Choice of algebra}
Algebra underlying product interaction has significant influence. For instance, if the underlying algebra for $3$ order product interaction  \eqref{eq:attention cubic}  is 
 direct sum 
$\mathcal{A}=\bigoplus_{i=1}^h \mathcal{A}^i$ , the resulting interaction is multi-head attention  with each component $\mathcal{A}^i$ as an head represents a distinct feature of the data.


The dynamical layer constructed in Section \ref{sec: Increasing self-interaction order} provides another example. Replacing Mamba's feature algebra \(\mathcal{A}_0\otimes\mathcal{A}\) with the simpler algebra \(\mathcal{A}\) reduces the dimensionality of the hidden state. As reported in Table~\ref{tab: lambdalambda}, this construction also achieves higher training and inference throughput. Thus, the choice of algebra influences both the form of the interaction and the computational requirements of the resulting architecture.

\paragraph{Structural operator}
Structural operators control how information is arranged, aggregated, or projected within a multiplication operator. For example, a causal mask can be implemented through causal reduction. Another example is the rank-one scalar projection $\mathcal{P}^0$ used to derive attention from a cubic product interaction.
A rank-$R$ projection $\mathcal{P}^R$ allows more information to be retained during projection. It also provides a construction analogous to multi-head interactions, in which different components can represent distinct features of the data.

\paragraph{Numerical evaluation of projection rank.} We examine the effect of projection rank on the copy task by evaluating the model $\mathcal{P}^c(\mathbf{F}(\mathcal{P}^R(X\mathcal{T}(X)))\mathcal{T}(X))$ for different values of $R$. The results are reported in Table~\ref{tab:copy2}.
\begin{table}[H]
\centering
\small 
\caption{Copy-task accuracy for different projection ranks $R$ at sequence length $192$.}
\label{tab:copy2}
\begin{tabular}{lccc}
\toprule
\textbf{Model} & $R=1$ & $R=2$ & $R=4$ \\
\midrule
length 192  & 0.3603 & 0.9173 & 0.9239 \\
\bottomrule
\end{tabular}
\end{table}
The accuracy increases substantially from $R=1$ to $R=2$, with a smaller improvement at $R=4$. These results indicate that the amount of information retained by the projection affects the performance of the resulting interaction. The projection rank therefore provides an additional architectural design parameter within the Product Interaction formalism.

\section{Conclusions}
We have developed the Product Interaction formalism as a common algebraic framework for representing and constructing neural network architectures. By expressing several established architectures through multiplication operators and their compositions, the framework identifies algebraic structures shared by models that are conventionally described in different terms.
The construction of Product Interactions involves two complementary design choices: the algebra in which data are represented and the composition of operators acting on that algebra. The choice of algebra determines the admissible interactions and their structural properties. In particular, symmetry constraints restrict the algebraic products, while simpler algebraic representations can reduce computational cost. Operator composition provides a second means of controlling the architecture. The notion of self-interaction order characterizes the dependence introduced through these compositions and offers a systematic way to construct higher-order interactions.
The numerical experiments illustrate the effects of these design choices on classification accuracy and computational efficiency. They also show that increasing self-interaction order does not necessarily improve performance monotonically. Determining which algebraic structures and operator compositions yield effective architectures, while controlling their computational cost, remains an important problem.
More broadly, the Product Interaction formalism provides a mathematical setting in which neural architectures can be studied through their algebraic structure rather than solely through their computational implementations. By making the structural choices underlying existing architectures explicit, it offers a foundation for moving beyond predominantly empirical architecture design toward a more systematic, mathematically grounded theory of neural network architectures.

\appendix

\section{Mamba and SSM}
\subsection{Background}\label{app:review mamba}
Let
$x(t)=(x_\alpha(t))\in\mathbb{R}^{d}$ denote the input. Classical state space model(SSM) process
$x(t)$ by mapping it to a higher dimension hidden state 
$h_{\alpha}(t)=(h_{\alpha i}(t))_{i}\in\mathbb{R}^{N}$ before project to output. The dynamical equation for SSM is 
\begin{equation}\label{eq:classical SSM}
\begin{cases}
\displaystyle 
\frac{d h_{\alpha i}}{dt}
   = \lambda_{\alpha i}h_{\alpha i}
     + B_{\alpha i} x_{\alpha},\\[0.6em]
\displaystyle 
y_{\alpha} 
= \sum_{i=1}^N
  C_{\alpha i} h_{\alpha i}.
\end{cases}
\end{equation}
Here we choose diagonal operator and omit $D$ part for simplicity.
%
Mamba models improve upon this by 
introducing the \emph{selection mechanism} by allowing both the input map
$B_{\alpha i}$ and the readout map $C_{\alpha i}$ to depend on the current input:
$
B(t)=W^{B}x(t),\quad C(t)=W^{C}x(t),
$
where $W^{B},W^{C}$ are linear maps.  The Mamba ODE is 
\begin{equation}\label{eq:mamba-classical}
\begin{cases}
\displaystyle 
\frac{d h_{\alpha i}}{dt}
   = \lambda_{\alpha i}h_{\alpha i}
     + \Big( \sum_{\beta} W^{B}_{i\beta}x_{\beta}\Big) x_{\alpha},\\[0.6em]
\displaystyle 
y_{\alpha} 
= \sum_{i}
  \Big(\sum_{\gamma} W^{C}_{i\gamma} x_{\gamma}\Big) h_{\alpha i}.
\end{cases}
\end{equation}
%
The discrete rule of Mamba ODE is 
\begin{equation*}
    h_{\alpha i}(t) =(1+\Delta_t^\alpha\lambda_{\alpha i})  h_{\alpha i}(t-1)+\Delta_t^\alpha \Big( \sum_{\beta} W^{B}_{i\beta}x_{\beta}(t)\Big) x_{\alpha}(t)
\end{equation*}
The selection mechanism also makes $\Delta_t^\alpha$ input dependent 
$$\Delta_t^\alpha=\sigma(\sum_\eta W^g_{\alpha\eta}x_\eta(t)+b_\alpha).$$
This input dependent $\Delta_t^\alpha$ plays the role of gating mechanism.
Approximation $\exp(\Delta_t^\alpha\lambda_{\alpha i})\approx1+\Delta_t^\alpha\lambda_{\alpha i}$ and $\Delta_t^\alpha\approx\frac{\exp(\Delta_t^\alpha\lambda_{\alpha i})-1}{\lambda_{\alpha i}}$ gives  Zero-Order Hold (ZOH) discretization in Mamba.
\subsection{Multiplication operator for SSM}
\label{app:SSM}
The expressions for $\mathcal{O}_B(X)$ and $\mathcal{O}_C(H)$ are
\begin{equation*}
    \mathcal{O}_B(X)=B\mathcal{T}(X)=\sum_{\alpha,j}(\sum_iB_{\alpha i}\lambda_{i0}^j)x_\alpha f_\alpha\otimes e_j,\quad  \mathcal{O}_C(H)=CH=\sum_{\alpha,k}(\sum_{i,j}C_{\alpha i}H_{\alpha j}\lambda_{ij}^k) f_\alpha\otimes e_k
\end{equation*}
Then the algebraic dynamical system \eqref{eq:SSM}
becomes
\begin{equation*}
\begin{cases}
\displaystyle 
\frac{dH_{\alpha j}}{dt}=\mathcal{W}(H) + \hat{B}_{\alpha j}x_\alpha,\\[0.4em]
\displaystyle 
y_{\alpha}
=\sum_{i,j}C_{\alpha i}H_{\alpha j}\lambda_{ij}^0
\end{cases}
\end{equation*}
where $\hat{B}_{\alpha j}=\sum_iB_{\alpha i}\lambda_{i0}^j$. This is exactly the dynamical
system 
\eqref{eq:classical SSM} for SSM.


\section{Equivariant Networks}

\subsection{Geometry background for TFN}
\label{app:geometry background for TFN}
Consider a manifold $M$ and trivial vector bundle $E_1=M\times V_1$. Suppose now we have a compact group $G$ which acts on $M$ together with a representation on $V_1$ denoted as
$g:M\to M$ and $\rho_1(g):V_1\to V_1$ for $g\in G$.

Then consider bundle homomorphism
\begin{tikzcd}
E_1 \arrow[r, "T_g"] \arrow[d, "\pi"'] 
  & E_1 \arrow[d, "\pi"] \\
M \arrow[r, "g"']                
  & M
\end{tikzcd}
where $T_g(p,v_p)=(gp,\rho_1(g)v_p)$ for $(p,v_p)\in M\times V_1$

Consider section of the vector bundle which in trivial vector bundle case can be seen as $\Gamma(E_1)=\{s:M\rightarrow V_1\}$.
Then bundle homomorphism $T_g$ induce transformation on $\Gamma(E_1)$ as
\begin{equation}
    (T_gs)(p)=\rho_1(g)s(g^{-1}p) \label{eq:genral transformation}.
\end{equation}

Consider another trivial vector bundle $E_2=M\times V_2$ with representation of $G$ defined on $V_2$ by $\rho_2(g)$. Then similarly there is also transformation $T_g$ defined on $\Gamma(E_2)$ 

Now consider an operator $\mathcal{L}:\Gamma(E_1)\rightarrow\Gamma(E_2)$, we say $\mathcal{L}$ is $G$ equivariant if 
\begin{equation}
    T_g\mathcal{L}(s)=\mathcal{L}(T_gs),\quad \forall s\in\Gamma(E_1) \quad \forall g \in G 
    \label{eq:general symmetry conditon}
\end{equation}

Let $M=\mathbb{R}^3$ and consider a trivial vector bundle
$E=M\times V$, where $V$ carries a (finite–dimensional) representation of
$SO(3)$, denoted $\rho(R):V\to V$.  


Decompose $V$ as a direct sum of irreducible $SO(3)$ representations,
$
   V = \bigoplus_{l\in\mathbb{N}} V_l,
   \;
   V_l = \mathrm{span}\{e^{l}_{m}\}_{m=-l,\dots,l},
$
so that for each $R\in SO(3)$
$
   \rho(R)e^{l}_{m}
    = \sum_{m'=-l}^{l} D^{(l)}_{m'm}(R)\,e^{l}_{m'},
$
where $D^{(l)}(R)$ is the Wigner–$D$ matrix.  Under this decomposition a field
$s$ can be written as
$
   s(\vec r)
   = \big( s^{l}_{m}(\vec r) \big)_{l,m},
   \;
   s^{l}_{m} : \mathbb{R}^{3}\to\mathbb{R}.
$

Follow the convention of TFN we consider action of rotation group \eqref{eq:genral transformation} as
\begin{equation}
    (T_{R}s)^{\,l}_{m}(\vec r)
      = \sum_{m'=-l}^{l} D^{(l)}_{m' m}(R)\,
        s^{l}_{m'}(R^{-1}\vec r).
    \label{eq:section-action-components}
\end{equation}

\subsection{Review of TFN and SE(3) attention}

Tensor Field Networks (TFNs) \cite{thomas2018tensor} process feature through following expression 
\begin{equation}\label{eq:standard TFN}
    f^{\text{out}}_{(l_o,m_o)}(\vec{r}_a)=\sum_bR^{(l_f,l_i)}(r_{ab})\sum_{m_f,m_i}C_{(l_f,m_f),(l_i,m_i)}^{(l_o,m_o)}Y_{(l_f,m_f)}(\hat{r}_{ab})f^{\text{in}}_{(l_i,m_i)}(\vec{r}_b)
\end{equation}
where $C_{(l_f,m_f),(l_i,m_i)}^{(l_o,m_o)}$ is Clebsch–Gordan coefficient, $Y_{(l_f,m_f)}$ is spherical harmonics, $r_{ab}=\| \vec{r}_a-\vec{r}_b \|$ and $\hat{r}_{ab}= \frac{\vec{r}_a-\vec{r}_b}{\| \vec{r}_a-\vec{r}_b \|}  $.

SE(3)-Transformers \cite{fuchs2020se3} combine TFN and attention in the following way: Key and value are obtained as edge information through
\begin{equation}\label{eq:k and v for se(3)}
k^{(l_o,m_o)}_{ab}/v^{(l_o,m_o)}_{ab}=\sum_{l_f,l_i}R_{l_o}^{(l_f,l_i)}(r_{ab})\sum_{m_f,m_i}C_{(l_f,m_f),(l_i,m_i)}^{(l_o,m_o)}Y_{(l_f,m_f)}(\hat{r}_{ab})f^{\text{in}}_{(l_i,m_i)}(\vec{r}_b)
\end{equation}
and query $q_a$ is obtained through TFN.
Then $SE(3)$ attention is updated through 
\begin{equation}\label{eq:standrad se(3) attention}
    \sum_{b\in\mathcal{N}_a-a}\alpha_{ab}v_{ab}, \qquad \alpha_{ab}=\frac{\exp{(q_a^Tk_{ab})}}{\sum_{b'\in\mathcal{N}_a-  a}\exp{(q_a^Tk_{ab'})}}
\end{equation}

Note that TFN and SE(3)-attention usually use real basis by a basis transformation.

\subsection{TFN as Multiplication operator}
\label{app:Symmetry principle for TFN}

We define transformation of $S\in\mathcal{B}\otimes\mathcal{B}\otimes\mathcal{A}$ follow geometry definition \eqref{eq:section-action-components}  and transfer it to representation of feature algebra.
\begin{align}
    \mathcal{T}_RS&=\sum_{a}\sum_l\sum_{m=-l}^l (T_Rs)_m^l(\vec{r_a})g_a\otimes g_0\otimes e_m^l \notag\\
    &=\sum_{a}\sum_l\sum_{m=-l}^l\sum_{m'=-l}^lD(R)_{m'm}^{(l)}s^l_{m'}(R^{-1}\vec{r_a}) g_a\otimes g_0\otimes e_m^l \notag\\
    &=\sum_{a}\sum_l\sum_{m'=-l}^ls^l_{m'}(R^{-1}\vec{r_a}) g_a\otimes g_0\otimes (\rho(R^{-1})^*e_{m'}^l) \label{eq:lift transformation}
\end{align}
Where $\rho(R^{-1})^*e_{m'}^l=\sum_{m=-l}^lD(R^{-1})^{(l)*}_{mm'}e_m^l=\sum_{m=-l}^lD(R)_{m'm}^{(l)}e_m^l$ and we use the unitary condition of $D(R)$.

\emph{Symmetry principle} \eqref{def:symmetry principle}
require that the lifted action $\mathcal{T}_R$ commutes with multiplication operator $\mathcal{O}_K$ 
Note that
\begin{align}
    \sum_{m=-l}^l K_m^l e_m^l&=\sum_mK_m^l[\rho(R^{-1})^*\rho(R)^*e_m^l]=\sum_{m}K_m^l[\sum_{m'}D(R)^{(l)*}_{m'm}\sum_nD(R^{-1})^{(l)*}_{nm'}e_n^l]\notag\\
&=\sum_{m'}(\sum_mD(R)^{(l)*}_{m'm}K_m^l)(\sum_nD(R^{-1})^{(l)*}_{nm'}e_n^l)=\sum_{m'}(\sum_mD(R^{-1})^{(l)}_{mm'}K_m^l)(\rho(R^{-1})^*e_{m'}^l) \label{eq: expression for K}
\end{align}

Then use \eqref{eq:lift transformation} and \eqref{eq: expression for K} we have
\begin{align*}
&\mathcal{O}_K(\mathcal{T}_R(S))=\mathcal{P}\big(K\mathcal{T}(\mathcal{T}_RS)\big)\\
&=\sum_{a,b}\sum_{l_1,m_1}\sum_{l_2,m_2}(\sum_{m=-l_1}^{l_1}D(R^{-1})_{mm_1}^{l_1}K_{m}^{l_1}(\vec{r_a}-\vec{r_b}))s_{m_2}^{l_2}(R^{-1}\vec{r_b})g_a\otimes g_0\otimes \rho(R^{-1})^*e_{m_1}^{l_1}\rho(R^{-1})^*e_{m_2}^{l_2}\\
&=\sum_{a,b}\sum_{l_1,m_1}\sum_{l_2,m_2}(\sum_{m=-l_1}^{l_1}D(R^{-1})_{mm_1}^{l_1}K_{m}^{l_1}(\vec{r_a}-\vec{r_b}))s_{m_2}^{l_2}(R^{-1}\vec{r_b})g_a\otimes g_0\otimes \rho(R^{-1})^*(e_{m_1}^{l_1}e_{m_2}^{l_2})
\end{align*}
where we have used the consist condition \eqref{eq:so3-compat}.
If we have
\begin{equation*}
    \sum_{m=-l_1}^{l_1}D(R^{-1})_{mm_1}^{l_1}K_{m}^{l_1}(\vec{r_a}-\vec{r_b})=K_{m_1}^{l_1}(R^{-1}\vec{r_a}-R^{-1}\vec{r_b})
\end{equation*}

Then
\begin{align*}
&\mathcal{O}_K(\mathcal{T}_R(S))
=\sum_{a,b}\sum_{l_1,m_1}\sum_{l_2,m_2}K_{m_1}^{l_1}(R^{-1}\vec{r_a}-R^{-1}\vec{r_b})s_{m_2}^{l_2}(R^{-1}\vec{r_b})g_a\otimes g_0\otimes \rho(R^{-1})^*(e_{m_1}^{l_1}e_{m_2}^{l_2})\\
&=\sum_{a,b}\sum_{l_1,m_1}\sum_{l_2,m_2}K_{m_1}^{l_1}(R^{-1}\vec{r_a}-\vec{r_b})s_{m_2}^{l_2}(\vec{r_b})g_a\otimes g_0\otimes \rho(R^{-1})^*(e_{m_1}^{l_1}e_{m_2}^{l_2})=\mathcal{T}_R(\mathcal{O}_K(S))
\end{align*}
where we used the fact that the sampling locations $\{\vec{r}_a\}$ are assumed to be rotation invariant.

So \eqref{eq:tfn-equiv-cond} holds if  if the kernel components obey
\begin{equation}
   \sum_{m=-l}^{l}
     D^{(l)}_{m m_1}(R^{-1})
     K^{\,l}_{m}(\vec r)
   =
   K^{\,l}_{m_1}(R^{-1}\vec r),
   \qquad\forall R\in SO(3).
   \label{eq:K-equiv}
\end{equation}
Solutions of \eqref{eq:K-equiv} are exactly of the form
$
K^{\,l}_{m}(\vec r)
   = R^{l}(\|\vec r\|)\,Y^{l}_{m}(\hat r),
$
with $Y^{l}_{m}$ spherical harmonics and $R^{l}$ an arbitrary radial profile.

\subsection{SE(3) attention as multi-level product interaction operator}
\label{app:SE(3) attention}

Just write $e_m^l$ as $e_n$ for simplicity $n=(l,m)$ and $e_0$ represents $e_{m=0}^{l=0}$. Then $S=\sum_{c,n}s_n(\vec{r}_c)g_c\otimes g_0\otimes e_n$, $W=\sum_{a,b,m}W_m(\vec{r}_{ab})g_a\otimes g_b\otimes e_m$.
The multiplication operator
\begin{align*}
\mathcal{O}_W(S)=W\mathcal{T}(S)&=\sum_{a,b}\sum_i(\sum_{n,m}C_{n,m}^is_n(\vec{r}_b)W_m(\vec{r}_{ab}))g_a\otimes g_b\otimes e_i =\sum_{a,b}\sum_i k_{a,b}^i g_a\otimes g_b\otimes e_i
\end{align*}
 gives the key used in SE(3)-attention \eqref{eq:k and v for se(3)}, simialr for value.
Then
\begin{align*}
\mathcal{O}_Q(K)=\mathcal{P}^0(QK)
&=\sum_{a,b}\sum_{j,i}C_{ji}^0q_j(\vec{r}_a)k_{ab}^ig_a\otimes g_b\otimes e_0=\sum_{a,b}\alpha_{ab}g_a\otimes g_b \otimes e_0
\end{align*}
Note that in real case $C_{lm,kn}^{00}$ gives inner product and in complex case we can apply a linear transformation to $q$ corresponding to complex conjugation to get inner product.
Thus this product gives us the attention weight used in SE(3)-attention.
Then
\begin{align*}
    &\mathcal{O}_A(V)=\mathcal{P}(AV)=\sum_{a,k}\sum_{\vec{r}_b\in \mathcal{N}_a}\mathbf{F}(\alpha_{ab})(\sum_{n,m}C_{mn}^kW_m(\vec{r}_{ab})s_n(\vec{r}_b))g_a\otimes g_0\otimes e_k
\end{align*}
gives SE(3) attention \eqref{eq:standrad se(3) attention} up to normalization of attention weight. 

\subsection{Symmetry principle}
\label{app: se(3) symmetry}
For edge element $E=\sum_{a,b}\sum_{l,m}E_m^l(\vec{r}_a,\vec{r}_b)g_a\otimes g_b\otimes e_m^l$, we define its transformation as 
\begin{equation}\label{eq: transformation of edge}
    \mathcal{T}_R(E)= \sum_{a,b}\sum_{l,m}E_m^l(R^{-1}\vec{r}_a,R^{-1}\vec{r}_b)g_a\otimes g_b\otimes \rho(R^{-1})^*e_m^l
\end{equation}
Note that
$$ \mathcal{T}_R(\mathcal{P}(E))=\sum_{a}\sum_{l,m}\big(\sum_bE_m^l(R^{-1}\vec{r}_a,\vec{r}_b)\big)g_a\otimes g_0\otimes \rho(R^{-1})^*e_m^l$$
and 
$$ \mathcal{P}(\mathcal{T}_R(E))=\sum_{a}\sum_{l,m}\big(\sum_bE_m^l(R^{-1}\vec{r}_a,R^{-1}\vec{r}_b)\big)g_a\otimes g_0\otimes \rho(R^{-1})^*e_m^l$$
Since sampling locations are rotation equivariant we have $\sum_bE_m^l(R^{-1}\vec{r}_a,\vec{r}_b)=\sum_bE_m^l(R^{-1}\vec{r}_a,R^{-1}\vec{r}_b)$, so $\mathcal{T}_R\circ\mathcal{P}=\mathcal{P}\circ\mathcal{T}_R$



\section*{Acknowledgments}
AIAR gratefully acknowledges the support of the Yau Mathematical
Sciences Center, Tsinghua University. This work is also
supported by the Tsinghua University Dushi Program. CWC is funded by the Swiss National Science Foundation (SNSF) under grant number 20HW-1 220785. 

\bibliographystyle{siamplain}
\bibliography{references}

\end{document}


\maketitle


\section{Experimental Details}
\label{app:experiment}

\subsection{Experimental details for Table 1 in main text}
For Vision Mamba baseline,
we adopt a multi-head bidirectional Vision Mamba classifier for CIFAR-10. Input images (32×32) are split into non-overlapping patches of size 4×4 via a convolutional patch embedding, yielding a sequence of 64 tokens with channel dimension $d_{\text{model}}=128$.
 A learnable positional encoding is added to the patch tokens before they are passed through a stack of 8 VMamba blocks. Each block consists of a residual connection around a pre-norm RMSNorm and a VMamba mixer. The mixer expands the input to an inner dimension $2\cdot d_{\text{model}}=256$
 via a linear projection, splits it into feature and gating branches, and applies a causal 1D convolution (kernel size 4). The core SSM dynamical layer is extended with $n_{\text{heads}}=4$ heads
( $d_\text{head}=64$, state dimension $d_{\text{hidden}}=16$, where the input-dependent parameters $\Delta$, $B$ and $C$
are produced by a linear layer and reshaped into multi-head tensors.  To capture bidirectional context, a mirrored backward branch with independent parameters processes the reversed sequence, and the two outputs are averaged. After the VMamba stack, we apply global average pooling, an RMSNorm, and a linear classification head for 10 classes.

For the new dynamical layer, we fix all the other part of VMamba architecture and only
 replace the mamba dynamical layer with our new dynamical layer.
We follow exactly the same hyperparameter as Vision Mamba. $B$ and $C$ are project to each head with dimension $d_{\text{hidden}}=16$ and $X$ is reshaped into each head with $d_\text{head}=64$, then $X$ and $B$ are multiplied to form hidden state $H$ with dimension $d_{\text{hidden}}=16$ in each head via a structure constant $\lambda$ learned for each head. The final output is get from product between $H$ and $C$ in each head via another shared structure constant $\lambda$.

Training is performed on CIFAR-10 with standard data augmentation. Inputs are normalized with dataset statistics. We use the AdamW optimizer with a learning rate of $1\times 10^{-3}$ and weight decay 0.05, together with a cosine annealing rschedule decaying to $1\times 10^{-6}$ over 300 epochs. The batch size is 128, and 10 percentage of the training set is held out for validation. The validation dataset is not used for training. We run 5 random seeds and select checkpoints with best validation accuracy for each seed and record corresponding test accuracy. Then we calculate mean and stand standard deviations. All training and inference experiments were conducted on an NVIDIA RTX 4090 GPU with 48 GB of VRAM and a 20‑core CPU. Training and inference speed measurements are reported based on runs performed on this instance under identical training configurations.

\subsection{Experimental details for Table 2 in main text}
Models with symmetry constraint corresponds to CNN model 
\begin{equation*}
  (\mathcal{O}_K(X))_{nm}
=
  \sum_{i,j} X_{ij}\, K_{\,i-n,\,j-m}
\end{equation*}
Models without symmetry constraint corresponds to model 
\begin{equation*}
    (\mathcal{O}_K(X))_{nm}
   =   \sum_{i,j}\sum_{k,l} X_{ij}K_{kl}\,
        \lambda^{\,n}_{k,i}\lambda^{\,m}_{l,j}
\end{equation*}
where $\lambda^{\,n}_{k,i}$ are learnable parameters without any constraint.

Models with regularized symmetry constraint corresponds to model 
\begin{equation}\label{eq:multi}
    (\mathcal{O}_K(X))_{nm}
   =   \sum_{i,j}\sum_{k,l} X_{ij}K_{kl}\,
        \lambda^{\,n}_{k,i}\lambda^{\,m}_{l,j}
\end{equation}
but with symmetry constraint $\lambda^{\,n}_{k,i+a}=\lambda^{\,n-a}_{k,i}$
enforced via regularization term
\begin{equation*}
    L_{\text{reg}}=\sum_{k,i,n,a}(\lambda^{\,n}_{k,i+a}-\lambda^{\,n-a}_{k,i})^2
\end{equation*}
in the loss.

The network architecture is LeNet for model with symmetry constraint. For models without symmetry constraint and with regularized symmetry constraint, we replace CNNs layer in LeNet by multiplication operator expression \eqref{eq:multi}.
The model is trained on 48GB vGPU with batch size $64$ for 100 epochs.

\subsection{Structure of Multi-head is critical for attention}
\label{app:multi-copy task}
We evaluate standard multi-head attention on hard KV CopyTask. We first provide the dataset description.

The \emph{Hard KV CopyTask} is a synthetic associative-retrieval benchmark designed to stress long-range memory under heavy distraction. Each training example is a length-$L$ token sequence $x \in \{0,1,\dots,n_{\text{classes}}\}^{L}$ constructed from three special symbols: a separator $\texttt{SEP}=0$, a key--value delimiter $\texttt{KV}=1$, and a query marker $\texttt{Q}=2$, while ordinary symbols occupy the range $\{3,\dots,n_{\text{classes}}\}$. The sequence begins with a \emph{store phase} that encodes a random dictionary of $M$ single-token key--value pairs with (approximately) unique keys, formatted as
$\texttt{SEP},\, k_1,\, \texttt{KV},\, v_1,\, \texttt{SEP},\, k_2,\, \texttt{KV},\, v_2,\, \texttt{SEP},\, \ldots,\, k_M,\, \texttt{KV},\, v_M,\, \texttt{SEP}$.
This is followed by a long \emph{distractor block} of randomly sampled ordinary symbols that occupies most of the remaining budget, forcing the model to preserve the stored mapping across a large irrelevant gap. Finally, the \emph{query phase} contains $Q_{\text{num}}$ retrieval requests, each encoded as a four-token pattern $\texttt{Q},\, k,\, \texttt{SEP},\, v$, where the queried key $k$ is sampled (with repetition allowed) from the stored keys and $v$ is the corresponding stored value. Supervision is posed as next-token prediction with targets $y[t]=x[t+1]$ for $t=0,\dots,L-2$, together with a binary mask $\text{mask}\in\{0,1\}^{L}$ that selects only the positions immediately preceding each query answer token; i.e., the loss is computed only when the model is required to predict the correct value associated with the queried key. This construction yields a challenging setting where correct performance requires robust long-context retention and key-conditioned retrieval despite substantial distractor interference.

Then we provide the model hyperparameter setting.
Unless otherwise stated, we use a Transformer-based sequence predictor with model dimension
$D = H \times d_h = 4 \times 32 = 128$. The backbone consists of $4$ stacked Transformer layers, each employing causal self-attention
with $4$ attention heads and a per-head dimension of $32$. Each layer includes a position-wise feed-forward network with hidden dimension $512$
(i.e., a $4\times$ expansion of the model dimension). Models are trained for $100$ epochs using a batch size of $64$. Optimization is performed using Adam.

More results are shown in Table~8
\begin{table}[H]
\centering
\small 
\caption{Copy task accuracy across different $R$ }
\begin{tabular}{lccc}
\toprule
\textbf{Model} & $R=1$ & $R=2$ & $R=4$ \\
\midrule
length 128  &0.3988  &0.9935  &0.9962 \\
length 192  & 0.3669 & 0.9125 & 0.9251 \\
length 256  & 0.3404 &0.8772  &0.8757  \\
length 320  & 0.3337 &0.3405  & 0.8166 \\
\bottomrule
\end{tabular}
\end{table}

\subsection{Experimental details for Table 3 in main text}
The model is 
\begin{equation*} 
    \mathcal{O}^{\text{cubic}}(X)=\mathcal{P}^c(\mathbf{F}\{\mathcal{P}^R(X\mathcal{T}(X))\})\mathcal{T}(X)
\end{equation*}
This gives attention type update rule
\begin{equation*}
    \sum_{l=1}^{k}\sum_{r=1}^RF(\sum_{\alpha,\beta}\lambda_{\alpha,\beta}^{r}
   x_{\alpha}^{(k)}x_{\beta}^{(l)})(
   \sum_{\eta}
   \lambda^{\theta}_{r,\eta}x^{(l)}_{\eta})
\end{equation*}
The learnable parameters are $R$ bilinear matrix $A_{\alpha,\beta}^{r}=\lambda_{\alpha,\beta}^{r}$ and $R$ value embedding matrix $W_{\eta,\theta}^r= \lambda^{\theta}_{r,\eta}$ for $r=1,..., R$.
The resulting model consists of 
$R$ parallel attention modules, analogous to a multi-head attention structure.

We evaluate this model on the same copy task as \ref{app:multi-copy task}. We use the same model setting as attention in \ref{app:multi-copy task}
with one head, but increase our new $R$ parameter. More results are shown in Table~9
\begin{table}[H]
\centering
\small 
\caption{Copy task accuracy across different $R$ }
\begin{tabular}{lccc}
\toprule
\textbf{Model} & $R=1$ & $R=2$ & $R=4$ \\
\midrule
length 128  & 0.4007 & 0.9884 & 0.9945 \\
length 192  & 0.3603 & 0.9173 & 0.9239 \\
length 256  & 0.3494 & 0.8627 & 0.8683 \\
length 320  & 0.3192 & 0.7259 & 0.8206 \\
\bottomrule
\end{tabular}
\end{table}

\section{Basic Representation Theory}\label{app:representation theory}
\subsection{Representation of Lie Group}
Consider vector space $\mathcal{A}$ and a Lie group $G$ which defines the transformation. Let $\mathfrak{g}$ be the Lie algebra (assume semisimple) of $G$ with basis $\{L^a\}$. Then we define a representation of $\mathfrak{g}$ on $\mathcal{A}$. This can be done to ask the chosen basis $\{e_n\}_n$ of $\mathcal{A}$ to be eigenvectors of Cartan elements of Lie algebra. Then $\mathcal{A}$ will be the direct sum of irreducible representation of $\mathfrak{g}$:
\begin{equation*}
    \mathcal{A}=\bigoplus_l \mathcal{A}_l, \quad \mathcal{A}_l=\texttt{span}\{e_{i_l},e_{i_l+1},...,e_{i_{l+1}-1}\},
\end{equation*}
where $\mathcal{A}_l$ is the highest weight $l$ representation. 
Then the representation of Lie group $G$ on $\mathcal{A}$ is induce by 
\begin{equation*}
    \rho(g)\phi=\exp(\sum_a\theta_aL^a)\phi,\quad g=\exp(\sum_a\theta_aL^a)\in G, \quad \phi\in \mathcal{A}. 
\end{equation*}

We consider explicit example of Lie group $SO(3)$ with Lie algebra $\mathfrak{so}(3)=\{M\in \mathbb{R}^{3\times3}| M^T=-M\}$.
A basis for $\mathfrak{so}(3)$ is chosen as 
$$
J_x = \begin{pmatrix}
0 & 0 & 0 \\
0 & 0 & -1 \\
0 & 1 & 0 \\
\end{pmatrix}, \quad
J_y = \begin{pmatrix}
0 & 0 & 1 \\
0 & 0 & 0 \\
-1 & 0 & 0 \\
\end{pmatrix}, \quad
J_z = \begin{pmatrix}
0 & -1 & 0 \\
1 & 0 & 0 \\
0 & 0 & 0 \\
\end{pmatrix}
$$
Take $L_x=iJ_x$, $L_y=iJ_y$, $L_z=iJ_z$ where $L_z$ is Cartan element. The highest weight $l$ representation is spanned by $\{e_m^l\}_{m=-l,...,l}$ with representation
\begin{equation*}
    L_ze_m^{l}=me_{m}^l,\quad L^2e_{m}^l=l(l+1)e_{m}^l.
\end{equation*}

For a rotation $R$ with rotation axis $\hat{n}=(n_x,n_y,n_z)$ and rotation angle $\theta$, the representation of this Lie group element is 
\begin{equation*}
\exp(-i\theta\hat{n}\cdot L)e_m^l=\sum_{m'=-l}^lD(R)_{m'm}^{(l)}e_{m'}^l
\end{equation*}
where $L=(L_x,L_y,L_z)$. $D(R)_{m'm}^{(l)}$ is called Wigner–D matrix.

\subsection{Consist condition between product and representation }
If we extend vector space $\mathcal{A}$ to algebra by defining a product, then there is further consist condition between product and representation that is product commutes with representation
\begin{equation}
    (\rho(g)e_n)(\rho(g)e_m)=\rho(g)(e_ne_m), \quad \forall g\in G.
    \label{eq:general condition between product and rep}
\end{equation}
If this condition is satisfied the product actually gives product representation of group and the structure constant will be determined by representation theory.

For $SO(3)$ case the product consist with representation is 
\begin{equation}\label{eq:cg-product}
    e_{m_1}^{l_1}e_{m_2}^{l_2}= \sum_{l}\sum_{m=-l}^lC^{lm}_{l_1m_1,l_2m_2}e_{m}^l.
\end{equation}
The structure constant $C^{lm}_{l_1m_1,l_2m_2}$  is called Clebsch–Gordan coefficient which is determined by representation theory.
Actually this algebra is realized by algebra of smooth functions on sphere $C^{\infty}(S^2)$ up to normalizing constant with basis being spherical harmonics $e_m^l=Y^l_{m}$.

\subsection{Convention for Representation}
\label{app:convention for rep}
Let group $G$ has a representation $\rho:G\rightarrow GL(V)$ on vector space $V$ with basis $\{e_n\}$. The column vector convention is 
\begin{equation*}
    \rho(g)e_n=\sum_kD(g)_{kn}e_k.
\end{equation*}
Then
\begin{equation*}
    \rho(g)(\sum_nc_ne_n)=\sum_k(\sum_nD(g)_{kn}c_n)e_k.
\end{equation*}
This is matrix acts on column vector. The group condition $\rho(g_2)\rho(g_1)e_n=\rho(g_2g_1)e_n$ gives
\begin{equation*}
    \sum_kD(g_2)_{lk}D(g_1)_{kn}=D(g_2g_1)_{ln}.
\end{equation*}

So if we use transpose of $D(g)$ to transform component $(c_n)$ then the product will be opposite and is representation of opposite group.


\bibliographystyle{siamplain}
\bibliography{references}

%% file: ex_shared.tex
\usepackage{lipsum}
\usepackage{amsfonts}
\usepackage{graphicx}
\usepackage{epstopdf}
\usepackage{algorithmic}
\ifpdf
  \DeclareGraphicsExtensions{.eps,.pdf,.png,.jpg}
\else
  \DeclareGraphicsExtensions{.eps}
\fi

\usepackage{enumitem}
\setlist[enumerate]{leftmargin=.5in}
\setlist[itemize]{leftmargin=.5in}

\newsiamremark{remark}{Remark}
\newsiamremark{hypothesis}{Hypothesis}
\crefname{hypothesis}{Hypothesis}{Hypotheses}
\newsiamthm{claim}{Claim}
\newsiamremark{fact}{Fact}
\crefname{fact}{Fact}{Facts}

\headers{Product Interaction: An Algebraic Formalism for Deep Learning Architectures}{Haonan Dong, Chun-Wun Cheng, and Angelica I. Aviles-Rivero}

\title{Product Interaction: An Algebraic Formalism for Deep Learning Architectures\thanks{Preprint}}

\author{Haonan Dong\thanks{Qiuzhen College, Tsinghua University 
  (\email{dhn23@mails.tsinghua.edu.cn}, ).}
\and Chun-Wun Cheng\thanks{University of Cambridge}
\and Angelica I Aviles-Rivero\thanks{YMSC, Tsinghua University
  (\email{aviles-rivero@tsinghua.edu.cn}, corresponding author).}}

\usepackage{amsopn}
